\documentclass[hyperref={colorlinks,citecolor=red,urlcolor=black,bookmarks=false,hypertexnames=true}]{article}

\usepackage{arxiv}

\usepackage[utf8]{inputenc} 
\usepackage[T1]{fontenc}    
\usepackage{url}            
\usepackage{booktabs}       
\usepackage{amsfonts}       
\usepackage{nicefrac}       
\usepackage{microtype}      
\usepackage{lipsum}         
\usepackage{graphicx}
\usepackage{subcaption}
\usepackage{doi}

\usepackage{colortbl}
\usepackage{listings}
\usepackage{bm}
\usepackage{physics}

\usepackage{listings}
\usepackage{xcolor}
\usepackage{wrapfig}

\usepackage{amsmath,accents}
\usepackage{cleveref}       

\usepackage[ruled,vlined]{algorithm2e}

\usepackage{appendix}
\usepackage{cite}

\usepackage{multirow}
\usepackage{array}

\usepackage{chngcntr}
\usepackage{epigraph}

\newcommand{\rasymbol}[1]{\accentset{\circ}{#1}}

\SetKwComment{Comment}{/* }{ */}

\newcommand{\ttl}{%
		Dynamical Systems Theory Behind\\
		a Hierarchical Reasoning Model
}

\title{\ttl}

\date{}

\author{ 
	\href{https://orcid.org/0000-0002-4930-1846}{\includegraphics[scale=0.06]{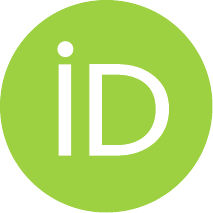}\hspace{1mm}Vasiliy A. Es'kin}$^{1,2}$\thanks{Corresponding author: Vasiliy Alekseevich Es’kin (\href{mailto:vasiliy.eskin@gmail.com}{vasiliy.eskin@gmail.com})},
	\href{https://orcid.org/0000-0002-0454-5249}{\includegraphics[scale=0.06]{orcid.pdf}\hspace{1mm}Mikhail E. Smorkalov}$^{2,3}$ \\
	\vspace{1mm}\\
	$^1$Department of Radiophysics, University of Nizhny Novgorod, Nizhny Novgorod 603950, Russia\\
	$^2$Huawei Nizhny Novgorod Research Center, Nizhny Novgorod 603006, Russia\\
	$^3$Skolkovo Institute of Science and Technology, Moscow 121205, Russia\\
	\vspace{1mm}\\
	\texttt{\href{mailto:vasiliy.eskin@gmail.com}{vasiliy.eskin@gmail.com}}, \texttt{smorkalovme@gmail.com}
}

\renewcommand{\headeright}{}
\renewcommand{\undertitle}{}
\renewcommand{\shorttitle}{A preprint}

\renewcommand{\vec}{\bf}

\usepackage{tikz}
\newcommand\tikznode[3][]%
{\tikz[remember picture,baseline=(#2.base)]
	\node[minimum size=0pt,inner sep=0pt,#1](#2){#3};%
}

\hypersetup{
pdftitle={Dynamical Systems Theory Behind a Hierarchical Reasoning Model},
pdfsubject={math.na, cs.AI, physics.comp-ph},
pdfauthor={Vasiliy A. Es'kin, Mikhail E. Smorkalov},
pdfkeywords={Dynamical Systems Theory,	Hierarchical Reasoning, Tiny Recursive Models, Neural ODEs, Stochastic Differential Equations, Algorithmic Reasoning, Parameter Efficiency, Contraction Mapping},
}

\begin{document}
\maketitle

\setlength{\epigraphwidth}{0.25\textwidth}
\epigraph{It was through hardship that the ape became human}{Folk wisdom}



	\vspace{-15mm}
\begin{center}
	\includegraphics[width=1.0\textwidth]{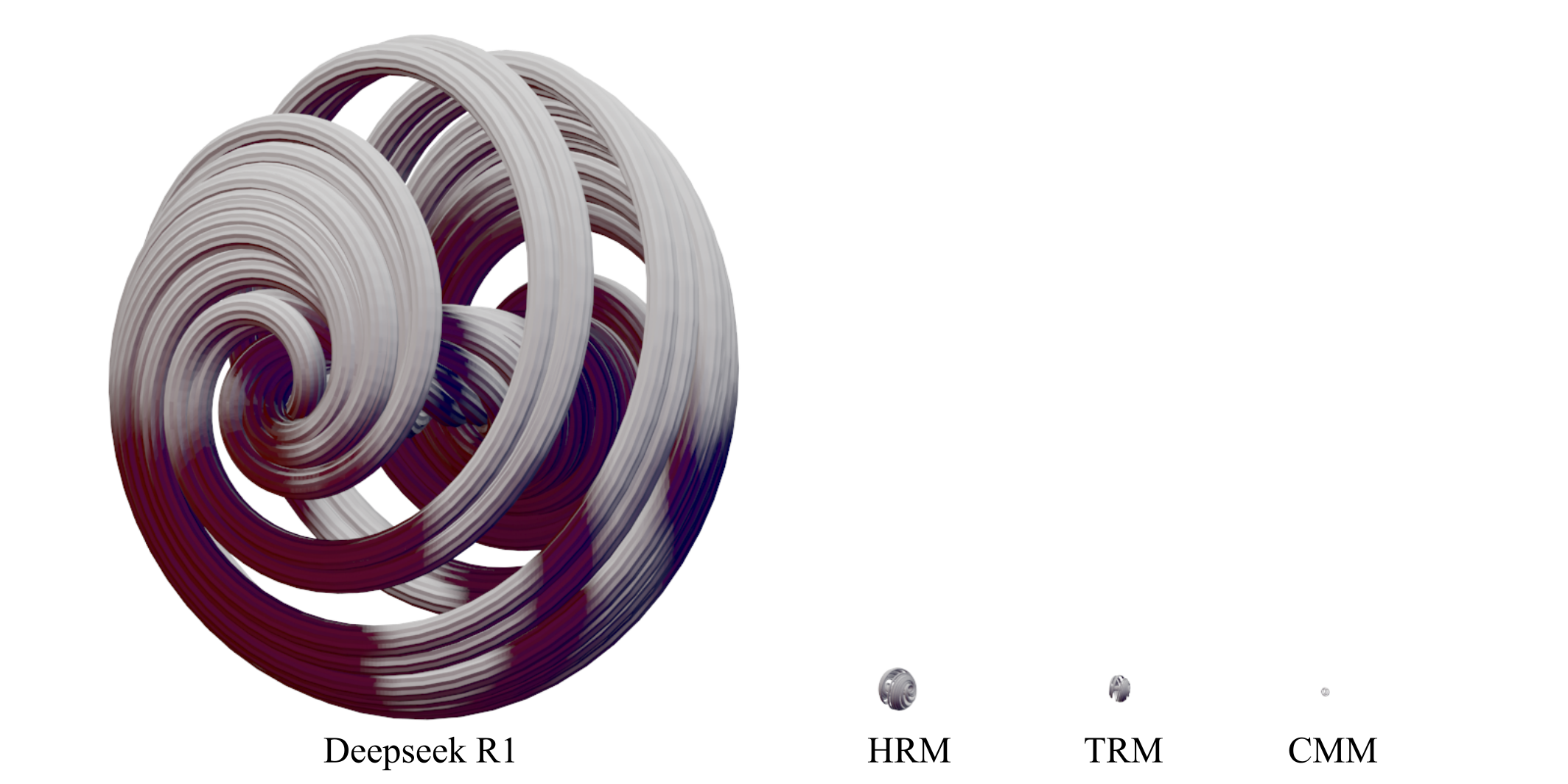}
	\captionof{figure}{Stylized visualizations of neural network models represented as volumetric Aizawa attractor trajectories. The number of model parameters is proportional to the volume of the sphere circumscribed around the corresponding attractor.}
	\vspace{1em}
\end{center}

\begin{abstract}
Current large language models (LLMs) primarily rely on linear sequence generation and massive parameter counts, yet they severely struggle with complex algorithmic reasoning. While recent reasoning architectures, such as the Hierarchical Reasoning Model (HRM) and Tiny Recursive Model (TRM), demonstrate that compact recursive networks can tackle these tasks, their training dynamics often lack rigorous mathematical guarantees, leading to instability and representational collapse. We propose the Contraction Mapping Model (CMM), a novel architecture that reformulates discrete recursive reasoning into continuous Neural Ordinary and Stochastic Differential Equations (NODEs/NSDEs). By explicitly enforcing the convergence of the latent phase point to a stable equilibrium state and mitigating feature collapse with a hyperspherical repulsion loss, the CMM provides a mathematically grounded and highly stable reasoning engine. On the Sudoku-Extreme benchmark, a 5M-parameter CMM achieves a state-of-the-art accuracy of 93.7\%, outperforming the 27M-parameter HRM (55.0\%) and 5M-parameter TRM (87.4\%). Remarkably, even when aggressively compressed to an ultra-tiny footprint of just 0.26M parameters, the CMM retains robust predictive power, achieving 85.4\% on Sudoku-Extreme and 82.2\% on the Maze benchmark. These results establish a new frontier for extreme parameter efficiency, proving that mathematically rigorous latent dynamics can effectively replace brute-force scaling in artificial reasoning.
\end{abstract}


\usetikzlibrary {arrows.meta,bending,positioning}

\section{Introduction}

The evolution of deep learning has been historically characterized by the systematic stacking of successive layers to enhance representational power, yet modern large language models (LLMs) based on the Transformer architecture remain paradoxically shallow in their fundamental computational structure~\cite{wang2025hierarchicalreasoningmodel}. Despite their success, these models operate primarily through a ``System 1'' modality, where the fixed-depth constraint of the Transformer block confines them to restricted computational complexity classes such as $AC^0$ or $TC^0$~\cite{Bylander91, merrill2025logicexpressinglogprecisiontransformers}. Specifically, models in $TC^0$ are limited to solving problems that can be decomposed into parallelizable, constant-depth circuits, effectively preventing standard Transformers from executing algorithms that require a number of sequential steps proportional to the input size~\cite{chiang2025transformersuniformtc0}. To circumvent these boundaries, the prevailing research paradigm has relied on Chain-of-Thought (CoT) prompting~\cite{wei2023chainofthoughtpromptingelicitsreasoning}, which externalizes reasoning into linear, token-level sequences. However, CoT is increasingly viewed as a functional crutch rather than an architectural solution; it suffers from high inference latency and extreme data requirements~\cite{wang2025hierarchicalreasoningmodel}. Inspired by the hierarchical organization of the human brain, recent advancements have introduced Hierarchical Reasoning Models (HRM) that separate high-level strategic planning from low-level procedural execution~\cite{ge2025hierarchicalreasoningmodelsperspectives}. A pivotal development in this space is the Tiny Recursive Model (TRM), which challenges traditional scaling laws by demonstrating that recursive weight-sharing can enable compact models to outperform much larger counterparts. Unlike standard Transformers, TRMs decouple computational depth from parameter count by iteratively passing latent representations through shared parameters for an arbitrary number of steps. This allows the model to expand its ``thinking time'' in the latent space, theoretically enabling it to transcend the $TC^0$ barrier within a compact, recursive architecture~\cite{jolicoeurmartineau2025morerecursivereasoningtiny}.

Despite the promising performance of hierarchical and recursive architectures on discrete reasoning benchmarks like ARC-AGI, several critical bottlenecks continue to impede the development of universal reasoning systems. A primary challenge lies in the efficiency of credit assignment within these multi-level structures. Current reinforcement learning (RL) methods often fail to distinguish between critical high-level planning decisions and routine execution steps, thereby diluting the optimization signal and slowing the discovery of effective latent strategies~\cite{wang2025emergenthierarchicalreasoningllms}. Furthermore, there is an ongoing debate regarding the necessity of explicit hierarchical modules; recent empirical evidence suggests that deep supervision and recursive refinement might be the primary drivers of performance, raising questions about whether current hierarchical separation truly captures the underlying structure of reasoning tasks~\cite{ge2025hierarchicalreasoningmodelsperspectives}. Training these models also faces systemic hurdles, including the high memory overhead of Backpropagation Through Time (BPTT) and the tendency of recurrent states to collapse into ``shortcut'' solutions that lack robustness~\cite{LILLICRAP201982, jolicoeurmartineau2025morerecursivereasoningtiny}. Perhaps most significantly, many current systems remain transductive, encoding specific task patterns into their weights rather than inducing generalizable algorithmic programs, which limits their adaptability to novel, out-of-distribution problems that require genuine ``program synthesis''~\cite{ARCFoundation2025, ge2025hierarchicalreasoningmodelsperspectives}. Furthermore, the emergence of TRMs raises a fundamental question regarding the minimal parameter set required to sustain complex reasoning without sacrificing the accuracy typically associated with much larger models. Consequently, a key area of interest lies in maximizing the reasoning precision of models within the compact parameter constraints established by state-of-the-art TRMs. These issues, ranging from the optimization of recursive depth to the pursuit of extreme parameter efficiency, constitute the frontier of current research in machine reasoning.

\begin{figure}[ht!]\centering
	\includegraphics[width=.43\textwidth]{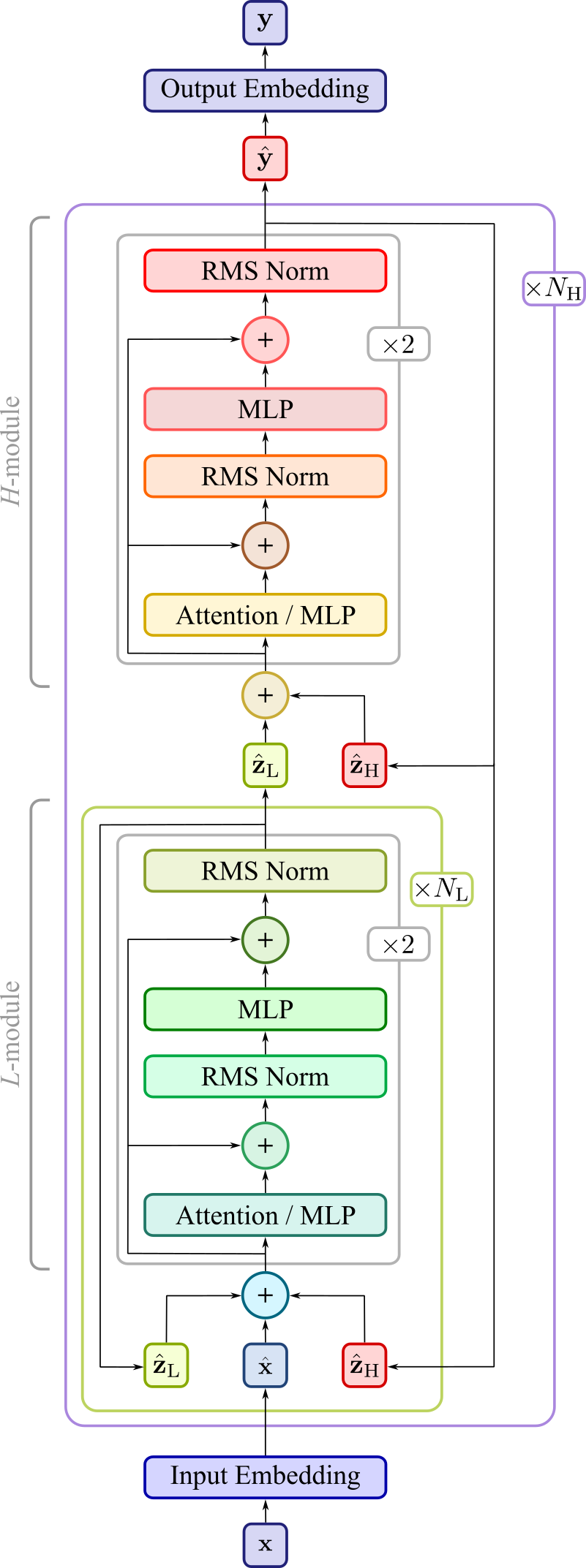}
	\caption{Architecture of hierarchical reasoning model.}\label{fig1}
\end{figure}

This work is devoted to enhancing the stability, generalization, and parameter efficiency of recursive reasoning models by analyzing them through the lens of continuous dynamical systems. We briefly list the main contributions of this paper:
\begin{enumerate}
	\item We reformulate the discrete recursive steps of hierarchical and tiny recursive architectures as continuous Neural Ordinary Differential Equations (NODEs) and Neural Stochastic Differential Equations (NSDEs), providing a rigorous mathematical framework for latent reasoning dynamics.
	\item We propose the Contraction Mapping Model (CMM), which enforces the convergence of the phase point of the system to a stable equilibrium state. This is achieved by introducing auxiliary loss terms derived directly from the equilibrium point conditions and Routh–Hurwitz stability criterion.
	\item We introduce a hyperspherical repulsion loss function designed to prevent representational collapse, ensuring a uniform, orthogonal, and highly expressive distribution of task features within the latent space.
	\item We develop computationally efficient polynomial approximations of the StableMax function (StableMax3 and StableMax5) and implement an adaptive loss balancing technique via Algebraic Gradient Normalization (AlgGradNorm) to stabilize the complex, multi-objective training process.
	\item We empirically demonstrate that CMMs achieve highly competitive accuracy on complex algorithmic reasoning benchmarks (Sudoku-Extreme and Maze) under extreme parameter constraints, successfully compressing the reasoning engine to just 0.26M parameters while matching or exceeding the performance of significantly larger baselines.
\end{enumerate}

The remainder of the paper is structured as follows. Section 2 reviews the problem statement and the current state of recursive reasoning architectures, specifically detailing the Hierarchical Reasoning Model (HRM) and Tiny Recursion Model (TRM) alongside their deep supervision training paradigms. In Section 3, we present our mathematical and architectural modifications, transitioning from discrete iterative equations to continuous NODEs and NSDEs, and define the proposed CMM with its auxiliary loss functions. Section 4 showcases the results of our extensive numerical experiments and ablation studies. Finally, concluding remarks are provided in Section 5.

\section{Current State and Statement of the Problem}
Consider a set of puzzles consisting of an input set $\left\{{\vec x}_{i}\right\}^{N_{p}}_{i=1}$ (questions) and an output set $\left\{{\vec y}_{i}\right\}^{N_{p}}_{i=1}$ (answers). Here, each input vector ${\vec{x}}_{i}$ corresponds to the output vector ${\vec{y}}_{i}$, and $N_{p}$ denotes the number of puzzles in the set. We will consider puzzles such as Sudoku-Extreme, Maze-Hard, ARC-AGI-1 and ARC-AGI-2 (see in~\cite{wang2025hierarchicalreasoningmodel}). It is necessary to find a mapping that translates the input data ${\vec x}_{i}$ into the output ${\vec y}_{i}$.

The process of solving the given problem can be considered as an operator $\mathcal{G}: \mathcal{X} \rightarrow \mathcal{Y}$ between spaces $\mathcal{X}$ and $\mathcal{Y}$. In the case of the puzzle problem, the space of questions $\mathcal{X}$ and the space of answers $\mathcal{Y}$ are finite spaces. We assume that there exists a solution operator $\mathcal{G}^{\dagger}$ for the given problem:
\begin{align}
	\mathcal{G}^{\dagger}: \mathcal{X} \rightarrow \mathcal{Y}.
\end{align} 
Our goal is to approximate $\mathcal{G}^\dagger$ by constructing a parametric map
\begin{align}
	\mathcal{G}_{\pmb \theta}: \mathcal{X} \rightarrow \mathcal{Y}, \quad {{\pmb \theta}} \in \Theta,
\end{align} 
for some finite-dimensional parameter space $\Theta$ by choosing ${{\pmb \theta}}^{\dagger} \in \Theta$ so that $\mathcal{G}_{{\pmb \theta}^{\dagger}} \approx \mathcal{G}^{\dagger}$. Such an operator can be represented by a large language model, or relatively small models such as hierarchical reasoning~\cite{wang2025hierarchicalreasoningmodel} and tiny recursive~\cite{jolicoeurmartineau2025morerecursivereasoningtiny} models.

\subsection{Hierarchical Reasoning Model}

Consider the Hierarchical Reasoning Model (HRM) as an approximator of the problem solving operator. The architecture of the HRM is shown on Figure~\ref{fig1}. The HRM model consists of the following learnable components: an input embedding ${\hat{\vec E}}_{\text{input}}(\cdot; {\bm \theta}_{\text{in}})$, a low-level recurrent module ${\hat{\vec F}}_{\rm L}(\cdot; {\bm \theta}_{\rm L})$ ($L$-module), a high-level recurrent module ${\hat{\vec F}}_{\rm H}(\cdot; {\bm \theta}_{\rm H})$ ($H$-module), and an output network ${\vec E}_{\text{output}}(\cdot; {\bm \theta}_{\text{out}})$. Dynamics of the model unfold over $N_{\rm H}$ high-level cycles of $N_{\rm L}$ low-level timesteps each. The discrete timesteps of a single forward pass are indexed by $i = 1, \dots, N_{\rm L} \times N_{\rm H}$. The modules ${\hat{\vec F}}_{\rm L}$ and ${\hat{\vec F}}_{\rm H}$ each produce hidden state tensors $\hat{\bf{z}}^{(i)}_{\rm L}$ for ${\hat{\vec F}}_{\rm L}$ and $\hat{\bf{z}}^{(i)}_{\rm H}$ for ${\hat{\vec F}}_{\rm H}$, which are initialized with the tensors $\hat{\bf{z}}^{(0)}_{\rm L}$ and $\hat{\bf{z}}^{(0)}_{\rm H}$, respectively.

The HRM maps an input vector $\vec x$ to an output prediction vector $\vec y$ as follows. First, the input $\vec x$ is projected into a hidden tensor of working representation $\hat{\vec x}$ by the input network:
\begin{align}\label{eq3}
	\hat{\vec x} =  {\hat{\vec E}}_{\text{input}}({\vec x}; {\bm \theta}_{\text{in}}).
\end{align} 

At each timestep $i$, the $L$-module updates its state conditioned on its own previous state $\hat{\bf{z}}^{(i-1)}_{\rm L}$, current state of the $H$-module $\hat{\bf{z}}^{(i-1)}_{\rm H}$ (which remains fixed throughout the cycle), and the input representation $\hat{\vec x}$. The $H$-module updates $\hat{\bf{z}}^{(i)}_{\rm H}$ only once per cycle (i.e., every $N_{\rm L}$ timesteps) using final state of the $L$-module at the end of that cycle:
\begin{align}
	& \hat{\bf{z}}^{(i)}_{\rm L} = {\hat{\vec F}}_{\rm L}\left(\hat{\bf{z}}^{(i-1)}_{\rm H} + \hat{\bf{z}}^{(i-1)}_{\rm L} + \hat{\vec x}; {\bm \theta}_{\rm L}\right),\label{eq4a} \\
	&\hat{\bf{z}}^{(i)}_{\rm H} = \begin{cases}
		{\hat{\vec F}}_{\rm H}(\hat{\bf{z}}^{(i-1)}_{\rm H} + \hat{\bf{z}}^{(i-1)}_{\rm L}; {\bm \theta}_{\rm H}) & \text{if}\quad i\equiv 0 \quad (\text{mod} \,\,\,N_{\rm L}),\\[6pt]
		\hat{\bf{z}}^{(i-1)}_{\rm H} & \text{otherwise}.
	\end{cases} \label{eq4b}
\end{align}

Finally, after $N_{\rm H}$ full cycles, a prediction $\vec y$ is extracted from the hidden state of the $H$-module:
\begin{align}\label{eq3}
	{\vec y} =  {\vec E}_{\text{output}}({\hat{\vec y}}; {\bm \theta}_{\text{out}}),
\end{align}
where $\hat{\vec y} = \hat{\bf{z}}^{(N_{\rm L}N_{\rm H})}_{\rm H}$.
This entire ${(N_{\rm L}N_{\rm H})}$--timestep process represents a single forward pass of the HRM.

\subsection{Tiny Recursion Model}
Tiny Recursive Models (TRMs), proposed in~\cite{jolicoeurmartineau2025morerecursivereasoningtiny}, represent a simple recursive reasoning approach using a single tiny network recursing on its latent reasoning feature and progressively improving its final answer. The $L$-module and $H$-module are represented by the same neural network (i.e., ${\bm \theta}_{\rm L} \equiv {\bm \theta}_{\rm H}$ and ${\hat{\vec F}}_{\rm L} \equiv {\hat{\vec F}}_{\rm H}$). This single module consists of two transformer layers (see Figure~\ref{fig1}) with a self-attention layer or an MLP-mixer (MLP)~\cite{tolstikhin2021mlpmixer}, which is applied on the sequence length. It significantly reduces the number of parameters by halving the number of layers and replacing the two networks with a single tiny network. Another feature of the TRM that mitigates overfitting and improves the generalizing abilities of the neural network is the use of an Exponential Moving Average (EMA). As a result of applying these approaches, the TRM significantly outperformed HRM in terms of accuracy.

\subsection{Training}

\begin{figure}[t!]\centering
	\includegraphics[width=\textwidth]{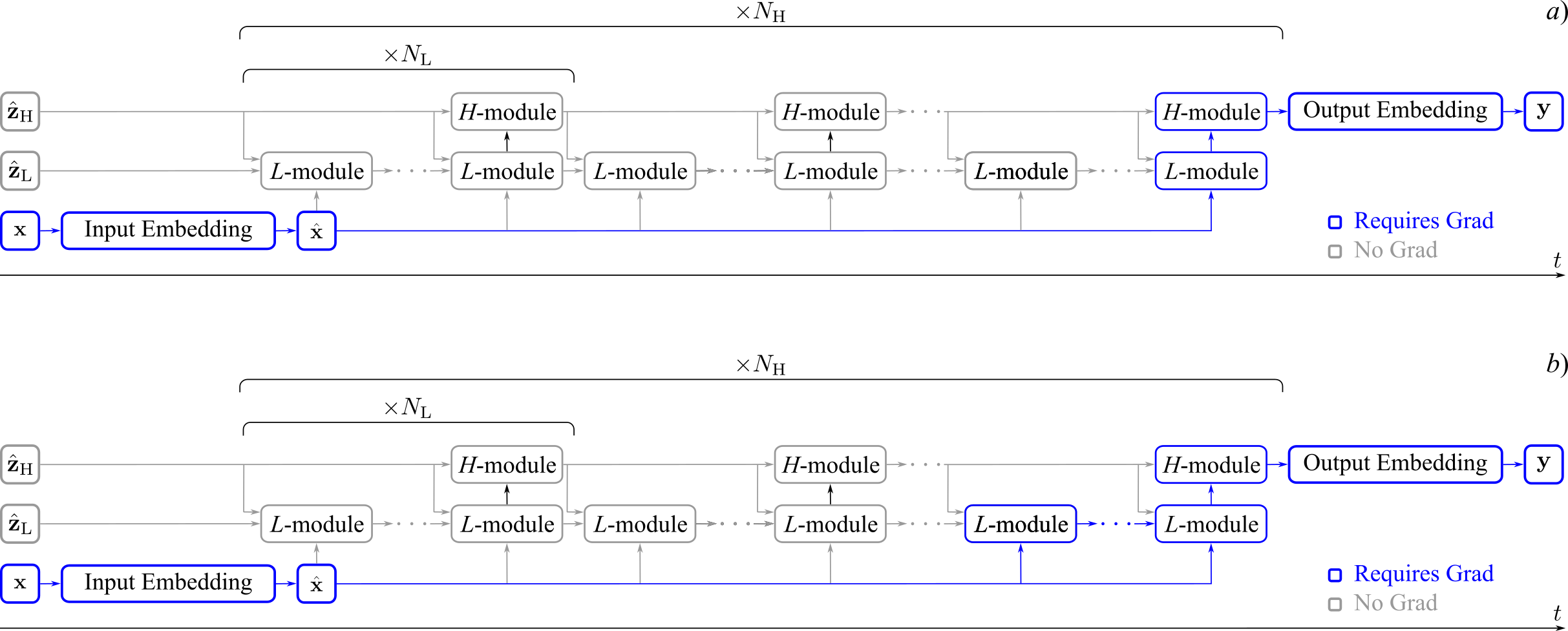}
	\caption{Diagram of training of HRM (a) and TRM (b).}\label{fig2}
\end{figure}

\begin{figure}[htbp]
	\centering
	{\scriptsize
		\lstset{
			basicstyle=\ttfamily\scriptsize,
			columns=fullflexible,
			aboveskip=2pt,
			belowskip=2pt,
			lineskip=-1pt
		}
	\begin{minipage}[t]{0.48\textwidth}
	\begin{lstlisting}[caption={Pseudocode of HRM}, label={lst1}]
def hrm(z, x, N_L=2, N_H=2): # hierarchical reasoning
	zH, zL = z
	with torch.no_grad():
		for i in range(N_L*N_H - 2):
			zL = L_net(zL, zH, x)
			if (i + 1) % N_H == 0:
				zH = H_net(zH, zL)
	# 1-step grad
	zL = L_net(zL, zH, x)
	zH = H_net(zH, zL)
	return (zH, zL), output_head(zH), Q_head(zH)
	
def ACT_halt(q, y_hat, y_true):
	target_halt = (y_hat == y_true)
	loss = 0.5*binary_cross_entropy(q[0], target_halt)
	return loss
	
def ACT_continue(q, last_step):
	if last_step:
		target_continue = sigmoid(q[0])
	else:
		target_continue = sigmoid(max(q[0], q[1]))
	loss = 0.5*binary_cross_entropy(q[1], target_continue)
	return loss
	
# Deep Supervision
for x_input, y_true in train_dataloader:
	z = z_init
	for step in range(N_supervision): # deep supervision
		x = input_embedding(x_input)
		z, y_pred, q = hrm(z, x)
		loss = softmax_cross_entropy(y_pred, y_true)
		# Adaptive computational time (ACT) using Q-learning
		loss += ACT_halt(q, y_pred, y_true)
		_, _, q_next = hrm(z, x) # extra forward pass
		loss += ACT_continue(q_next, step == N_sup - 1)
		z = z.detach()
		loss.backward()
		opt.step()
		opt.zero_grad()
		if q[0] > q[1]: # early-stopping
			break
	\end{lstlisting}
	\end{minipage}
	\hfill
	\begin{minipage}[t]{0.50\textwidth}
	\begin{lstlisting}[caption={Pseudocode of TRM}, label={lst2}]
def latent_recursion(x, y, z, N_L=6):
	for i in range(N_L): # latent reasoning
		z = net(x, y, z)
	y = net(y, z) # refine output answer
	return y, z
	
def deep_recursion(x, y, z, N_L=6, N_H=3):
	# recursing N_H-1 times to improve y and z (no gradients needed)
	with torch.no_grad():
		for j in range(N_H-1):
			y, z = latent_recursion(x, y, z, N_L)
	# recursing once to improve y and z
	y, z = latent_recursion(x, y, z, N_L)
	return (y.detach(), z.detach()), output_head(y), Q_head(y)
	
# Deep Supervision
for x_input, y_true in train_dataloader:
	y, z = y_init, z_init
	for step in range(N_supervision):
		x = input_embedding(x_input)
		(y, z), y_hat, q_hat = deep_recursion(x, y, z)
		loss = softmax_cross_entropy(y_hat, y_true)
		loss += binary_cross_entropy(q_hat, (y_hat == y_true))
		loss.backward()
		opt.step()
		opt.zero_grad()
		if q_hat > 0: # early-stopping
			break
	\end{lstlisting}
	\end{minipage}
}	
	\caption{Pseudocode of training of the HRM (left) and TRM (right).}
	\label{fig3}
\end{figure}

The authors of HRM proposed a one-step approximation of the HRM gradient, using the gradient of the last state of each module and treating other states as constants (see Figure~\ref{fig2}a). The authors of the TRM modified this training step by using the last state of the $H$-module for gradient calculation, as well as the last states of the $L$-modules after the previous update of the state of the $H$-module (see Figure~\ref{fig2}b).

\subsubsection{Deep supervision for the HRM}
For a given data sample ($\vec x$, $\vec y$), multiple forward passes of the HRM model are run, each of which is referred to as a segment. For each segment $m\in \left\{1,\dots,N_s\right\}$, let $\hat{\vec z}^{(m)} = \left(\hat{\bf{z}}^{(m N_{\rm L}N_{\rm H})}_{\rm H} , \hat{\bf{z}}^{(m N_{\rm L}N_{\rm L})}_{\rm L}\right)$ represent the hidden state at the end of segment $m$. At each segment $m$, a deep supervision step of training is applied as follows:
\begin{align}
	\left(\hat{\vec z}^{(m)}, {\vec y}^{(m)}\right)& \leftarrow \text{HRM}\left(\rasymbol{\hat{\vec z}}^{(m-1)}, {\vec x};{\bm \theta}\right),\notag\\
	\mathcal{L}^{(m)}_{\rm LM}({\bm \theta})& \leftarrow {\text{Loss}}\left({\vec y}^{(m)}, {\vec y}\right), \notag\\
	{\bm \theta} &\leftarrow {\text{OptimizationStep}}\left({\bm \theta}, \nabla_{\bm \theta} \mathcal{L}^{(m)}_{\rm LM}({\bm \theta}) \right).
\end{align}
Here, $\text{HRM}$ is the HRM model, $\hat{\vec z}^{(m-1)}$ is the hidden state on the previous segment $m-1$, ${\vec y}^{(m)}$ is the output vector associated with the current hidden state (at segment $m$) $\hat{\vec z}^{(m)}$, ${\bm \theta}$ is the set of all trainable parameters of HRM (${\bm \theta} = \left\{{\bm \theta}_{\text{in}}, {\bm \theta}_{\text{H}}, {\bm \theta}_{\text{L}}, {\bm \theta}_{\text{out}}\right\}$). Here and below Ra symbol over variables indicates that the computed values ($\rasymbol{\hat{\vec z}}^{(m-1)}$) are detached from the computational graph (this operation corresponds to \verb*|detach| in PyTorch, and \verb*|lax.stop_gradient| in JAX). The critical aspect of this process is that the hidden state, $\rasymbol{\hat{\vec z}}^{(m)}$, is ``detached'' from the computation graph before being used as the input state for the next segment. Consequently, gradients from segment $m + 1$ do not propagate back through segment $m$, effectively creating a one-step approximation of the gradient of the recursive deep supervision process~\cite{bai2022deepequilibriumopticalflow, ramzi2023shinesharinginverseestimate}. This approach provides more frequent feedback to the $H$-module and acts as a regularization method, demonstrating better empirical performance and improved stability in deep equilibrium models compared to more complicated, Jacobian-based regularization methods~\cite{bai2022deepequilibriumopticalflow, bai2021stabilizingequilibriummodelsjacobian}. Figure~\ref{fig3} shows the pseudocode for training the HRM using deep supervision.

\subsubsection{Adaptive computational time (ACT)} \label{secACT}
The authors of the HRM incorporated an adaptive halting strategy into the HRM. This integration leverages deep supervision and uses the $Q$-learning algorithm to adaptively determine the number of segments. A $Q$-head uses the final state of the $H$-module to predict the $Q$-values $\hat{Q}^{(m)} = \left(\hat{Q}^{(m)}_{\rm halt}, \hat{Q}^{(m)}_{\rm continue}\right)$ of the ``halt'' and ``continue'' actions:
\begin{align}
	& \hat{Q}^{(m)} =\sigma \left({\bm \theta}^{\rm T}_{Q} \hat{\bf{z}}^{(m N_{\rm L}N_{\rm H})}_{\rm H} \right),
\end{align}
where $\sigma$ is a sigmoid function, ${\bm \theta}^{\rm T}_{Q}$ are trainable parameters. Let $M_{\rm max}$ and $M_{\rm min}$ are the maximum (a fixed hyperparameter) and minimum (a random variable) numbers of segments, respectively. The value of $M_{\rm min}$ is determined randomly: with a probability $\varepsilon$, it is chosen uniformly from the set $\{2, \dots, M_{\rm min}\}$, and with a probability $1-\varepsilon$, it is set to $1$. The halt action is taken under two conditions: when the number of segments exceeds the maximum threshold $M_{\rm max}$, or when the estimated value of the halt $\hat{Q}^{(m)}_{\rm halt}$ is greater than the estimated value for continuing $\hat{Q}^{(m)}_{\rm continue}$, and the number of segments has reached at least the minimum threshold $M_{\rm min}$.

The $Q$-head is updated through a $Q$-learning algorithm, which is defined by Markov Decision Process (MDP). The state of the MDP at segment $m$ is $\hat{\vec z}^{(m)}$, and the action space is $\{\text{halt}, \text{continue}\}$. Choosing the action ``halt'' terminates the episode and returns a binary reward indicating prediction correctness, i.e., ${\vec 1}\,\{{\vec y}^{(m)} = {\vec y}\}$. Choosing ``continue'' yields a reward of $0$ and the state transitions to $\hat{\vec z}^{(m+1)}$. Thus, the $Q$-learning targets for the two actions $\hat{G}^{(m)} = \left(\hat{G}^{(m)}_{\rm halt}, \hat{G}^{(m)}_{\rm continue}\right)$ are given by
\begin{align}
	\hat{G}^{(m)}_{\rm halt} & = {\vec 1}\,\{{\vec y}^{(m)} = {\vec y}\},\notag \\
	\hat{G}^{(m)}_{\rm continue} & = \begin{cases}
	\hat{Q}^{(m+1)}_{\rm halt}, & \text{if}\quad m\geq M_{\rm max} \\
	\max\left(\hat{Q}^{(m+1)}_{\rm halt},\hat{Q}^{(m+1)}_{\rm continue}\right), & \text{otherwise}.
	\end{cases}
\end{align}
The overall loss for each supervision segment combines both the $Q$-head loss and the sequence-to-sequence loss:
\begin{align}
	&\mathcal{L}^{(m)}_{\rm ACT}({\bm \theta}) := \mathcal{L}^{(m)}_{\rm LM}({\bm \theta}) + \mathcal{L}^{(m)}_{\rm BCE}({\bm \theta}),
\end{align}
where
\begin{align}
	& \mathcal{L}^{(m)}_{\rm BCE}({\bm \theta}) := \text{BinaryCrossEntropy}\left(\hat{Q}^{(m)}, \hat{G}^{(m)}\right).
\end{align}

Minimizing this loss allows for both accurate predictions and near-optimal stopping decisions. The ``halt'' action ends the training loop. In practice, sequences are processed in batches, which can be easily handled by replacing any halted sample in a batch with a new sample from the data loader.

The optimization problem can be defined as follows:
\begin{equation}
	{\bm \theta}^* = {\arg}\,\underset{{\bm \theta}}{\min} \,\mathcal{L}^{(m)}_{\rm ACT}({\bm \theta}),\label{eq9}
\end{equation}
where ${\bm \theta}^*$ are optimal parameters of the neural networks which minimize the discrepancy between the exact solution ${\vec y}$ and the approximate one ${\vec y}^{(m)}$.

\subsubsection{Training of TRM}
The training of the TRM is similar to the training of the HRM described above. There are two main differences. 

The first difference is the reduction of the need for the expensive second forward pass. The ACT in HRM through Q-learning requires two forward passes, which slows down training. The authors of the TRM proposed a simple solution, which is to eliminate the continue loss (from the Q-learning) and only learn a halting probability through a Binary-Cross-Entropy loss of having reached the correct solution. By removing the continue loss, the authors eliminated the need for the expensive second forward pass, while still being able to determine when to halt with relatively good accuracy. 

The second difference is the use of an Exponential Moving Average (EMA).
To reduce the problem of overfitting in the HRM due to small datasets and to improve stability, the authors integrated Exponential Moving Average (EMA) of the weights, a common technique in GANs and diffusion models to improve stability. It is found that it prevents sharp collapse and leads to higher generalization.

\section{Modifications of the HRM}
\subsection{From Discrete Equations to Neural Ordinary Differential Equations}\label{fromDiscrete}
We need the continuous counterparts of equations~(\ref{eq4a}) and (\ref{eq4b}) for further analysis.
Next, we find the differential equations that generate these difference equations. Let the discrete variables be related to continuous time as
\begin{align}
	\hat{\bf{z}}^{(i)}_{\rm L} = \hat{\bf{z}}_{\rm L}(t_i), \quad \hat{\bf{z}}^{(i)}_{\rm H} = \hat{\bf{z}}_{\rm H}(t_i), \quad t_i = i \Delta t.
\end{align}
The time derivative is approximated by the forward difference:
\begin{align}
	& \left.\frac{{\rm d} z}{{\rm d} t}\right|_{t=t_i} \approx \frac{{{z}}(t_i) - {{z}}(t_{i-1})}{\Delta t} = \frac{{{z}}^{(i)} - {{z}}^{(i-1)}}{\Delta t}.
\end{align}
From (\ref{eq4a}) we obtain the corresponding differential equation
\begin{align}
	\frac{\partial \hat{\bf{z}}_{\rm L}(t)}{\partial t} = \frac{1}{\Delta t}\left[{\hat{\vec F}}_{\rm L}\left(\hat{\bf{z}}_{\rm H}(t) + \hat{\bf{z}}_{\rm L}(t) + \hat{\vec x}; {\bm \theta}_{\rm L}\right) - \hat{\bf{z}}_{\rm L}(t)\right].
\end{align}
The equation (\ref{eq4b}) corresponds to an impulse differential equation
\begin{align}
	\frac{\partial \hat{\bf{z}}_{\rm H}(t)}{\partial t} = \frac{1}{\Delta t}\left[{\hat{\vec F}}_{\rm H}\left(\hat{\bf{z}}_{\rm H}(t) + \hat{\bf{z}}_{\rm L}(t); {\bm \theta}_{\rm H}\right) - \hat{\bf{z}}_{\rm H}(t)\right] \sum_{j\in \mathbb{Z}}\delta\left(t - j N_{\rm L} \Delta t\right),
\end{align}
where $\delta$ is the Dirac delta function. Averaging over the period $N_{\rm L} \Delta t$ leads to the following equation:
\begin{align}
	\frac{\partial \hat{\bf{z}}_{\rm H}(t)}{\partial t} = \frac{1}{N_{\rm L} \Delta t}\left[{\hat{\vec F}}_{\rm H}\left(\hat{\bf{z}}_{\rm H}(t) + \hat{\bf{z}}_{\rm L}(t); {\bm \theta}_{\rm H}\right) - \hat{\bf{z}}_{\rm H}(t)\right].
\end{align}

Assuming that $\Delta t = 1$, we get the following system of differential equations:
\begin{align}
	& \frac{\partial \hat{\bf{z}}_{\rm L}(t)}{\partial t} = {\hat{\vec F}}_{\rm L}\left(\hat{\bf{z}}_{\rm H}(t) + \hat{\bf{z}}_{\rm L}(t) + \hat{\vec x}; {\bm \theta}_{\rm L}\right) - \hat{\bf{z}}_{\rm L}(t),\label{eq17} \\
	&\frac{\partial \hat{\bf{z}}_{\rm H}(t)}{\partial t} = \frac{1}{N_{\rm L}}\left[{\hat{\vec F}}_{\rm H}(\hat{\bf{z}}_{\rm H}(t) + \hat{\bf{z}}_{\rm L}(t); {\bm \theta}_{\rm H}) - \hat{\bf{z}}_{\rm H}(t)\right].\label{eq18}
\end{align}

Note that the system of equations (\ref{eq4a}) and (\ref{eq4b}) represents a discrete Euler integration scheme with step size $\Delta t = 1$ applied to the continuous-time system of equations (\ref{eq17})--(\ref{eq18}). Let us rewrite this continuous system in the following more compact form
\begin{align}
	\frac{\partial \hat{\bf{z}}(t)}{\partial t} = \hat{\vec F}\left(\hat{\bf{z}}(t), \hat{\vec x}; {\bm \theta}\right) - \hat{\bf{z}}(t), \label{eq19}
\end{align}
where $\hat{\bf{z}}$ denotes either $\hat{\bf{z}}_{\rm L}$ or $\hat{\bf{z}}_{\rm H}$, and $\hat{\vec F}$ stands for ${\hat{\vec F}}_{\rm L}$ or ${\hat{\vec F}}_{\rm H}$, respectively.
For concreteness, we consider $t \in [0,T]$, where $T$ is the end time point. According to the problem formulation, the final value of $\hat{\bf{z}}_{\rm H}$ at $t=T$ must be equal to $\hat{y}$, while the initial values of $\hat{\bf{z}}_{\rm L}$ and $\hat{\bf{z}}_{\rm H}$ are not specified and can therefore be chosen arbitrarily. We assume the initial condition $\hat{\bf{z}}_{\rm H}(0) = \hat{\bf{x}}$. Thus, the initial and final values of $\hat{\bf{z}}_{\rm H}$ can be expressed as follows:
\begin{align}
	& \hat{\bf{z}}_{\rm H}(0) = \hat{\bf{x}},\quad \hat{\bf{z}}_{\rm H}(T) = \hat{\bf{y}}. \label{eq20}
\end{align}

The system (\ref{eq19}) belongs to the class of Neural Ordinary Differential Equations (NODEs)~\cite{chen2019NODE}, which, subject to conditions (\ref{eq20}), can be solved using standard ODE solvers. The objective is to optimize the neural networks within (\ref{eq19}) such that the resulting trajectory satisfies both the differential equations and the specified initial and final conditions.

It is important to highlight a key technical detail: the term $\hat{\vec x}$ (the input tensor of the puzzle) in the system of equations (\ref{eq19}) acts as a system parameter. This implies that for each specific puzzle, the model solves a distinct system of equations parameterized by the input tensor $\hat{\vec x}$. Consequently, the problem can be reformulated with a more relaxed initial condition compared to (\ref{eq20}): the objective is to find the neural network parameters ${\bm \theta}^*$ of $\hat{\vec F}$ such that, for any initial conditions of $\hat{\bf{z}}_{\rm L}$ and $\hat{\bf{z}}_{\rm H}$, the solution to the system of equations (\ref{eq19}) parameterized by the input tensor $\hat{\vec x}$ yields the puzzle solution $\hat{\bf{y}}$ at time $T$.


\subsection{Modifications of the training and model}

Note that a neural network can achieve the desired result (solution $y$) at the end of one of the first segments of adaptive computational time (Section~\ref{secACT}). Consequently, adaptive computational time in the following segments must not cause the state to diverge from this solution. That is, such a state of the system should be stable. In the theory of dynamical systems, such a state corresponds to stable equilibrium state or an attractive set (attractor)~\cite{andronov1974qualitative,Shilnikov2001methods}. Hereafter, we will call it a stable equilibrium point, encompassing the notion of attractive sets.

In this work, we show that the stability and convergence of recursive reasoning can be significantly improved by framing models like HRM and TRM as contraction mapping systems. To fully leverage this mathematical framework, we propose the Contraction Mapping Model (CMM). Unlike previous empirical methods, CMM enforces the contraction mapping property through specific architectural modifications and introduces auxiliary loss terms associated with equilibrium points. This provides a mathematically rigorous foundation for convergence of the model to a unique fixed point. Hereafter, for brevity, we will denote the CMM trained to solve the NODEs~(\ref{eq19}) subject to the conditions~(\ref{eq20}) as $\text{CMM}^{\text{NODE}}_D$, where $D$ is the dimensionality of the hidden state.

\subsubsection{Equilibrium points} 
The equilibrium points of dynamic system (\ref{eq19}) are determined from the following equation 
\begin{align}
	& \hat{\bf{z}}^*(t) = \hat{\vec F}\left(\hat{\bf{z}}^*(t), \hat{\vec x}; {\bm \theta}\right).\label{eq21}
\end{align}
Hereafter, by equilibrium points, we mean sets that can be attractive (stable equilibrium points) or repulsive (unstable equilibrium points). Let the function $\hat{\vec F}$ be expanded into a Taylor series in some neighborhood of the equilibrium state $\hat{\bf{z}}^*$. Then the system (\ref{eq19}) in this neighborhood can be rewritten as
\begin{align}
	& \frac{\partial {{z}}_i(t)}{\partial t} = \left[\sum\limits_{k=1}^{K} \left.\frac{\partial {F}_i\left(\hat{\bf{z}}(t), \hat{\vec x}; {\bm \theta}\right)}{\partial z_k}\right|_{\hat{\bf{z}}=\hat{\bf{z}}^*} - 1\right] \left({{z}}_i(t) - {{z}}^*_i\right).\label{eq22}
\end{align}
Here, $z_i$ and $F_i$ are the $i$th components of the tensor $\hat{\bf{z}}$ and the tensor function $\hat{\vec F}$ ($i=1,...,K$). In equation (\ref{eq22}), we neglected the nonlinear terms above the first order of smallness.

The characteristic equation of the linearized system (\ref{eq22}) is as follows
\begin{align}
D(\lambda) = 
\begin{vmatrix}
	J_{11} - \lambda & J_{12} & \cdots & J_{1K} \\
	J_{21} & J_{22} - \lambda & \cdots & J_{2K} \\
	\vdots & \vdots & \ddots & \vdots \\
	J_{K1} & J_{K2} & \cdots & J_{KK} - \lambda
\end{vmatrix}
= 0, \label{eq23}
\end{align}
where $J_{ik}=\dfrac{\partial \hat{\vec F}_i\left(\hat{\bf{z}}(t), \hat{\vec x}; {\bm \theta}\right)}{\partial z_k} - \delta_{ik}$, $\delta_{ik}$ is the Kronecker symbol. Expanding the determinant in (\ref{eq23}) yields a characteristic equation whose left side is a $K$-th degree polynomial with respect to $(\lambda)$:
\begin{align}
a_0\lambda^K + a_1\lambda^{K-1} + \dots + a_{K-1}\lambda + a_K = 0, \label{eq24}
\end{align}
where $a_0, a_1,\dots, a_K$ are real coefficients depending on the system parameters ($a_0 = (-1)^{K}$).

\textbf{Criterion of Routh-Hurwitz.} Let $a_0 > 0$. We form the determinants:
\begin{align}
\Delta_1 = a_1; \quad \Delta_2 = 
\begin{vmatrix}
	a_1 & a_0 \\
	a_3 & a_2
\end{vmatrix}; \quad \Delta_3 = 
\begin{vmatrix}
	a_1 & a_0 & 0 \\
	a_3 & a_2 & a_1 \\
	a_5 & a_4 & a_3
\end{vmatrix}; \quad \dots
\end{align}
\begin{align}
\Delta_K = 
\begin{vmatrix}
	a_1 & a_0 & 0 & 0 & \cdots & 0 \\
	a_3 & a_2 & a_1 & a_0 & \cdots & 0 \\
	a_5 & a_4 & a_3 & a_2 & \cdots & 0 \\
	\vdots & \vdots & \vdots & \vdots & \ddots & \vdots \\
	a_{2K-1} & a_{2K-2} & a_{2K-3} & a_{2K-4} & \cdots & a_K
\end{vmatrix}
= a_K \Delta_{K-1},
\end{align}
where $a_j = 0$, if $j > K$.

For all roots of equation (\ref{eq24}) (with real values $a_j$, $j=0, 1, \dots, K$, and given $a_0 > 0$) to have negative real parts, it is necessary and sufficient that the following inequalities hold:
\begin{align}\Delta_1 > 0, \quad \Delta_2 > 0, \quad \dots, \quad \Delta_{K-1} > 0; \quad a_K > 0.
\end{align}
As is known, the equilibrium state of a linear time-invariant system is considered asymptotically stable (i.e., the system returns to this state after small perturbations) if all roots ($\lambda_i$) of its characteristic equation lie in the open left half-plane of the complex plane. This means that the real part ($\text{Re}$) of every root must be strictly negative: $\text{Re}(\lambda_i) < 0$ for all $i$. Thus, if the conditions of the Routh-Hurwitz criterion are met, it guarantees that all roots of the characteristic equation lie in the left half-plane, and, consequently, the equilibrium state of the dynamic system is stable.

We should note that while satisfying all $K$ conditions of the Routh-Hurwitz criterion is strictly necessary and sufficient for asymptotic stability, computing all $K$ determinants is computationally intractable for high-dimensional latent spaces (e.g., $K \in \{128, 512\}$) within the inner loop of deep learning optimization. However, if the first condition $\Delta_1 > 0$ is not met, the system cannot be stable. Therefore, we adopt a computationally efficient relaxation by explicitly enforcing this first necessary condition. Geometrically, enforcing a positive $\Delta_1$ (which corresponds to a negative trace of the linearized system matrix) ensures that the local phase volume contracts around the equilibrium point. For the given problem, we have the following equation for $\Delta_1$:
\begin{align}\label{eq28}
	\Delta_1 = \frac{(-1)^{K-1}}{a_0}\left(\sum\limits_{i=1}^{K}J_{ii} - K\right)= K -\sum\limits_{i=1}^{K}J_{ii}.
\end{align}

\textbf{Additional Loss Terms.} Based on equations (\ref{eq21}) and (\ref{eq28}), we formulate the following additional loss terms for the values $\hat{\vec z}^{(m)}_{\rm H}$ of $m$-th segment. Equation (\ref{eq21}) yields the term:
\begin{align}\label{eq29}
	\mathcal{L}^{(m)}_{\rm equil}({\bm \theta})& := \left[\rasymbol{\hat{\bf{z}}}^{(n)}_{\rm H} -
		\hat{\vec F}_{\rm H}(\hat{\bf{z}}^{(n)}_{\rm H} + \hat{\bf{z}}^{(n)}_{\rm L}; {\bm \theta}_{\rm H})  \right]^2,
\end{align}
where $n= m N_{\rm L}N_{\rm H}$.


Equation (\ref{eq28}) yields the following auxiliary loss term. Rather than enforcing the computationally intractable full criterion, this term acts as a feasible regularizer corresponding to the first necessary condition (the trace penalty) derived from the Routh-Hurwitz criterion. This encourages the formation of a stable equilibrium point at $\hat{\vec z}^{*}_{\rm H}=\hat{\vec y}$:
\begin{align}\label{eq30}
	\mathcal{L}^{(m)}_{\rm RH\,stable}({\bm \theta})& := \left[{\rm ReLU}
	\left(\frac{1}{K}\sum\limits_{i=1}^{K}J_{ii}-1\right)  \right]^2.
\end{align}
Similarly, we introduce a corresponding loss term that encourages an unstable equilibrium point (a repeller) at $\hat{\vec z}^{*}_{\rm H}=\hat{\vec x}$:
\begin{align}\label{eq31}
	\mathcal{L}^{(m)}_{\rm RH\, unstable}({\bm \theta})& := \left[{\rm ReLU}
	\left(1 - \frac{1}{K}\sum\limits_{i=1}^{K}J_{ii}\right)  \right]^2.
\end{align}
Hereafter, for brevity, we refer to the loss terms derived from this necessary condition simply as ``RH stable'' and ``RH unstable''.

\textbf{Discussion on changing integral trajectories during the training process.}

\begin{figure}[ht!]\centering
	\includegraphics[width=0.9\textwidth]{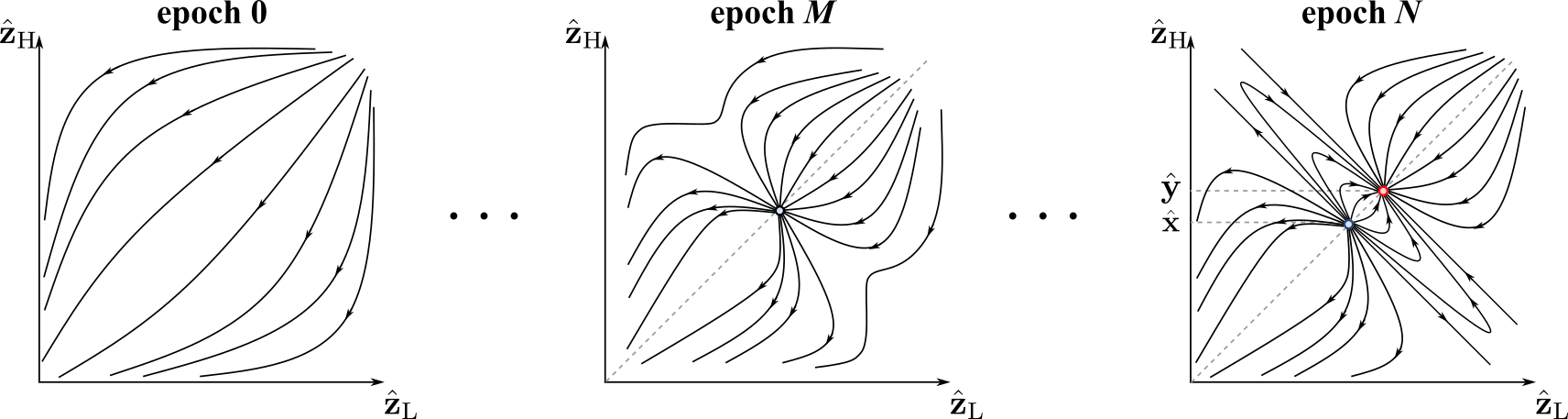}
	\caption{Evolution of the phase portrait of the dynamical system (\ref{eq19}), (\ref{eq20}) during neural network training.}\label{fig3_2}
\end{figure}

During the training of the neural network $\hat{\vec F}$, and taking into account the aforementioned auxiliary loss terms, the phase portrait of the dynamical system evolves such that a stable equilibrium point $\hat{\vec z}^{*}_{\rm H}=\hat{\vec y}$ emerges, toward which all integral trajectories converge. Figure~\ref{fig3_2} presents a sketch of the evolution of the phase portrait for the dynamical system (\ref{eq19})--(\ref{eq20}).

\begin{figure}[ht!]\centering
	\includegraphics[width=0.35\textwidth]{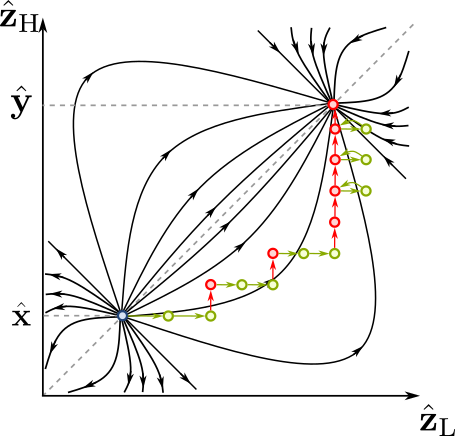}
	\caption{Integration of the system of differential equations (\ref{eq19})--(\ref{eq20}) under the trained neural network, which determines the landscape of phase space. Green and red arrows correspond to the integration stages using the $L$- and $H$-modules, respectively.}\label{fig3_3}
\end{figure}

It should be noted that as the system of differential equations is integrated and the phase point approaches the stable equilibrium point $\hat{\vec z}^{*}_{\rm H}=\hat{\vec y}$, the time-dependent behaviour of $\hat{\vec z}_{\rm L}$ (across integration steps) should become oscillatory. This is due to the fact that upon reaching the equilibrium, if $\hat{\vec z}_{\rm L}$ deviates from its equilibrium value at step $i$, it must take a value at step $i+1$ that is close to its value at step $i-1$. If this condition is violated, the phase point leaves the neighborhood of the equilibrium $\hat{\vec z}^{*}_{\rm H}=\hat{\vec y}$, and the solution of NODEs fails to satisfy the second condition (\ref{eq20}). This likely explains why an even number of steps for the $L$-module yields the best results for the TRM~\cite{jolicoeurmartineau2025morerecursivereasoningtiny}. An illustration of such behaviour with an even number of steps for the $L$-module is presented in Fig.~\ref{fig3_3}.

\subsubsection{Repulsion Loss Term}
	
To prevent representation collapse, ensure a uniform distribution of puzzle features within the latent space, and enhance the expressive power of the neural network, a repulsion-based loss function defined on a unit hypersphere is employed.  

Let the hidden representation (denoted generically as $\hat{\mathbf{z}}_{\mathrm{H}}$, $\hat{\mathbf{z}}_{\mathrm{L}}$, or $\hat{\mathbf{x}}$) be a tensor of dimensions $B \times S \times D$, where $B$ is the batch size, $S$ is the sequence length (the dimension of the input message), and $D$ is the hidden dimension. 

Initially, the feature tensor of the $i$-th sample in the batch is flattened into a single one-dimensional vector $\mathbf{z}_i \in \mathbb{R}^{M}$, where $M = S \times D$. These flattened vectors are subsequently projected onto the surface of a unit hypersphere via $L_2$-normalization:
\begin{equation}\label{eq32}
	\mathbf{u}_i = \frac{\mathbf{z}_i}{\|\mathbf{z}_i\|_2}, \quad i \in \{1, \dots, B\}.
\end{equation}
Following normalization, the Gram matrix (cosine similarity matrix) is constructed for all pairs of normalized vectors. To minimize the correlation between distinct samples, a penalty is imposed on the squared off-diagonal elements of this matrix. The final repulsion loss, $\mathcal{L}_{\text{rep}}$, is defined as the average of the squared inner products for all pairs $i \neq j$:
\begin{equation}\label{eq33}
	\mathcal{L}_{\text{rep}} := \frac{1}{B(B-1)} \sum_{i=1}^{B} \sum_{j \neq i} \langle \mathbf{u}_i, \mathbf{u}_j \rangle^2.
\end{equation}

\begin{figure}[ht!]\centering
	\includegraphics[width=\textwidth]{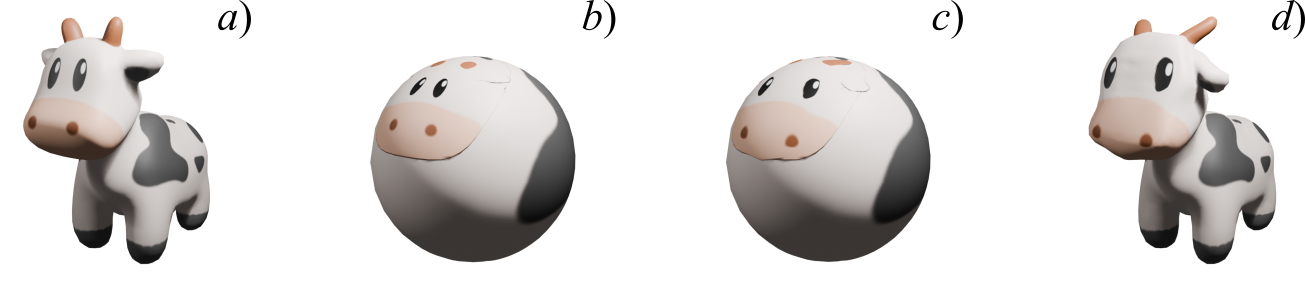}
	\caption{Illustration to repulsion loss term. (a) is initial hidden representation, (b) is projection to the surface of a unit hypersphere, (c) is projection after using repulsion loss term, (d) is hidden representation after using repulsion loss term.}\label{fig3_1}
\end{figure}

Squaring the inner product ensures that both parallel and anti-parallel representations are equally penalized, thereby driving the latent vectors toward mutual orthogonality. The diagonal elements are explicitly excluded from the summation to avoid penalizing the inherent self-similarity of the features. An illustration of the data transformation stages and their behaviour under the influence of this loss term is presented in Fig.~\ref{fig3_1}. Here, the cow's horns, eyes, and nostrils serve as the endpoints of the feature vector (``spot cow'' is taken from the repository associated with~\cite{crane2013robust}).

\subsubsection{Modifications of StableMax}

The authors of the HRM and TRM trained their models with the StableMax loss instead of Softmax for improved stability~\cite{prieto2025grokkingedgenumericalstability}:
\begin{align}
	& {\rm StableMax}(x_i) := \dfrac{s(x_i)}{\sum\limits_{j} s(x_j)},\label{eq34}\\
	& s(x) := \left\{\begin{array}{l}
		1 + x,          \quad x \ge 0,\\[2mm]
		\dfrac{1}{1-x}, \quad x < 0.
	\end{array} \right.\label{eq35}
\end{align}

To maintain the stability inherent in StableMax and increase the accuracy inherent in Softmax, the following modifications of StableMax were used in training: StableMax3 and StableMax5. StableMax3 is defined as follows
\begin{align}
	& {\rm StableMax3}(x_i) := \dfrac{s_3(x_i)}{\sum\limits_{j} s_3(x_j)},\label{eq36}\\
	& s_3(x) := \left\{\begin{array}{l}
		1 + x (1 + 0.5 x (1 + x / 3)),          \quad x \ge 0,\\[2mm]
		\dfrac{1}{1 - x (1 - 0.5 x (1 - x / 3))}, \quad x < 0.
	\end{array} \right.\label{eq37}
\end{align}

StableMax5 is defined as follows
\begin{align}
	& {\rm StableMax5}(x_i) := \dfrac{s_5(x_i)}{\sum\limits_{j} s_5(x_j)},\label{eq38}\\
	& s_5(x) := \left\{\begin{array}{l}
		1 + x (1 + 0.5 x (1 + x (1 + x (1 + x / 5) / 4) / 3)),          \quad x \ge 0,\\[2mm]
		\dfrac{1}{1 - x (1 - 0.5 x (1 - x (1 - x (1 - x / 5) / 4) / 3))}, \quad x < 0.
	\end{array} \right.\label{eq39}
\end{align}
In Eqns.~(\ref{eq37}) and~(\ref{eq39}), $1 + x (1 + 0.5 x (1 + x / 3))$ and $1 + x (1 + 0.5 x (1 + x (1 + x (1 + x / 5) / 4) / 3))$ are computationally efficient Taylor series expansions of $\exp(x)$ up to and including the third and fifth powers, respectively, implemented via the Horner method for polynomial evaluation. This motivates the use of the numbers 3 and 5 in the names of the proposed methods, StableMax3 and StableMax5.

\subsection{Neural Stochastic Differential Equations}

The training process descriptions for HRM and TRM (see Figure~\ref{fig3} and \cite{wang2025hierarchicalreasoningmodel, jolicoeurmartineau2025morerecursivereasoningtiny}) state that the training procedure includes inner loops over \verb*|N_supervision| steps, which implement the ACT scheme. However, the official code repositories for these models \cite{wang2025hierarchicalreasoningmodel, jolicoeurmartineau2025morerecursivereasoningtiny} omit this explicit inner loop. Instead, they employ only an outer loop over batches via a \verb*|train_dataloader|, while the current ACT segment is tracked internally within the model. Consequently, \textbf{the batch iterations and ACT segment iterations are fused}. This setup implies that the initial hidden states $\hat{\vec z}^{(m)}_i$ for a given segment $m$ and batch $i$ are the final hidden states $\hat{\vec z}^{(m-1)}_{i-1}$ from the previous segment and batch. 

\begin{figure}[ht!]\centering
	\includegraphics[width=\textwidth]{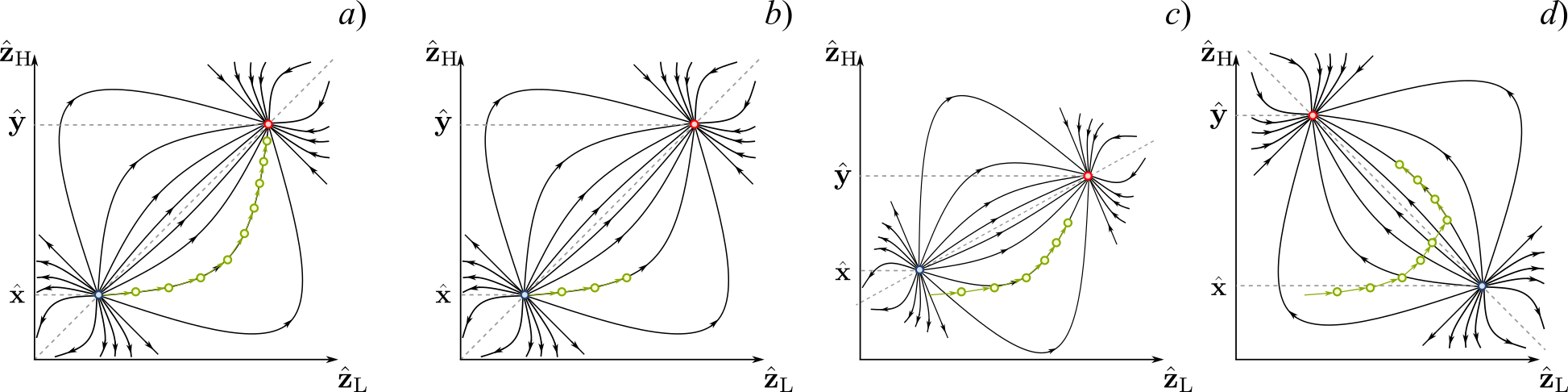}
	\caption{(a) represents the integration of the system of differential equations \eqref{eq19} during training according to the descriptions in \cite{wang2025hierarchicalreasoningmodel, jolicoeurmartineau2025morerecursivereasoningtiny}; (b), (c), and (d) represent the integration of the system of differential equations \eqref{eq19} during training according to the official code for these papers. Panels (b), (c), and (d) correspond to batches $i$, $i+1$, and $i+2$, respectively.}\label{fig7}
\end{figure}

This training implementation causes the phase point (representing the system states $\hat{\mathbf{z}}_{\mathrm{L}}$ and $\hat{\mathbf{z}}_{\mathrm{H}}$ at a given moment) to jump from one integral trajectory to another, rather than moving smoothly along a single trajectory toward a stable equilibrium point. A demonstration of the difference in behaviour between the integration described in the papers and the integration implemented in the code is shown in Fig.~\ref{fig7}. Fig.~\ref{fig7}(a) illustrates the behaviour of the phase point under the integration procedure when an explicit loop over $N_{\text{supervision}}$ is present. In the case where the $N_{\text{supervision}}$ loop and the dataset batch loop are fused, the integration trajectory of the phase point changes direction in accordance with the shifting stable equilibrium point (see Fig.~\ref{fig7}(b)--(d)).

The behaviour described above is equivalent to a situation where, for a given puzzle, the initial values of $\hat{\vec z}$ for the $m$-th segment take on random values every $N_{\rm L} {\times} N_{\rm H}$ integration steps. Considering this circumstance and the remark at the end of section \ref{fromDiscrete}, we have modified the NODEs (\ref{eq19}) into the following neural stochastic differential equations (NSDEs) \cite{liu2019neuralsdestabilizingneural,shen2025neuralsdesunifiedapproach}:
\begin{align}
	{{\rm d} \hat{\bf{z}}(t)} = \left[\hat{\vec F}\left(\hat{\bf{z}}(t), \hat{\vec x}; {\bm \theta}\right) - \hat{\bf{z}}(t)\right]{{\rm d} t} + G\left(\hat{\bf{z}}(t), t \right) {{\rm d} \hat{\vec W}(t)}, \label{eq40}
\end{align}
where $W(t)$ is a standard Wiener process (Brownian motion)~\cite{Oksendal2003}, which is a continuous-time stochastic process such that $\hat{\vec W}(t+s) - \hat{\vec W}(s)$ follows a Gaussian distribution $\mathcal{N}(0, s \hat{\vec I})$. Here, $G\left(\hat{\bf{z}}(t), t \right)$ represents the noise injection (which can be either additive $G\left(t \right)$ or multiplicative $G\left(\hat{\bf{z}}(t), t \right)$). The discrete version of equation~\ref{eq40} for the additive noise can be written as
\begin{align}
	\hat{\bf{z}}^{(i)} = \hat{\vec F}\left(\hat{\bf{z}}^{(i-1)}, \hat{\vec x}; {\bm \theta}\right) + \sigma \hat{\bm{\zeta}}^{(i)}, \label{eq41}
\end{align}
where $\hat{\bm{\zeta}}^{(i)} \sim \mathcal{N}\left(0, \hat{\vec I}\right)$ is standard Gaussian noise, and $\sigma$ is the magnitude of the stochastic component. The dimensions of $\hat{\bm{\zeta}}^{(i)}$ and $\hat{\bf{z}}^{(i)}$ coincide.
The discrete version of equation~\ref{eq40} for the multiplicative noise can be written as
\begin{align}
	\hat{\bf{z}}^{(i)} = \hat{\vec F}\left(\hat{\bf{z}}^{(i-1)}, \hat{\vec x}; {\bm \theta}\right)\odot \left(\hat{\vec I} + \sigma \hat{\bm{\zeta}}^{(i)}\right). \label{eq42}
\end{align}
Hereafter, for brevity, we will denote the CMM trained to solve the NSDEs~(\ref{eq40}) as $\text{CMM}^{\text{NSDE}}_D$.

It is evident that the injection of noise improves both the accuracy and the generalization capabilities of the model \cite{liu2019neuralsdestabilizingneural, shen2025neuralsdesunifiedapproach}. This stochasticity creates a manifold of latent states from which the phase point is forced to converge toward the stable equilibrium. As specified in equations (\ref{eq41}) and (\ref{eq42}), the magnitude $\sigma$ acts as a regularization hyperparameter that directly determines the spatial extent of this latent state distribution.

Without the explicit noise component, the model would likely be trained to reach the target state only along narrow, specific trajectories, leading to rapid overfitting. In this context, the inherent ``jumps'' between batches in the fused iteration cycle, combined with the explicitly added noise $\sigma \hat{\bm{\zeta}}^{(i)}$, act synergistically to prevent representational collapse. This combination promotes the formation of a more robust and smooth manifold for the latent dynamics, ultimately enhancing the performance of the solver on unseen data.

\subsection{Weights of Loss Terms}

Taking into account the additional loss components in the optimization problem (\ref{eq9}) (which was formulated for the HRM and TRM), for the CMM, it is necessary to replace $\mathcal{L}^{(m)}_{\rm ACT}({\bm \theta})$ with the sum of all loss terms. Naturally, each component of the loss function must be included with a certain weight. Thus, the final loss is expressed as follows:
\begin{align}\label{eq44}
	&\mathcal{L}^{(m)}_{\rm Total}({\bm \theta}) := \sum\limits_{\substack{n \in \text{terms}}} \lambda_n \mathcal{L}^{(m)}_{n}({\bm \theta}),
\end{align}
where $\lambda_n$ is the weight of the $n$-th loss term, ``$\text{terms}$'' is the set $\{\text{LM}, \text{BCE}, \text{``rep x''}, \text{``rep z''}, \text{``equil x''}, \text{``equil z''}, \text{``RH stable z''},$ $\text{``RH unstable x''}\}$,  ``x'' and ``z'' in the names of the terms denote the loss terms for $\hat{\vec x}$ and $\hat{\vec z}_{\rm H}$, respectively (for example, ``rep x'' and ``rep z'' correspond to repulsion loss for $\hat{\vec x}$ and $\hat{\vec z}_{\rm H}$).

The choice of weights $\lambda_n$ can be performed empirically to balance the contribution of each loss term. The task-specific terms (LM and BCE) require larger magnitudes compared to the auxiliary stability terms (repulsion and equilibrium) to ensure that the model prioritizes the correct puzzle solution. The final values were kept constant across a portion of our experiments to maintain consistency.

\subsubsection{Adaptive Loss Balancing via Algebraic Gradient Normalization (AlgGradNorm)}
Experience in balancing loss terms for training physics-informed neural networks (PINNs)~\cite{Eskin2024,ESKIN2025114085,es2025separable} suggests that the most effective approach is the use of automatic balancing techniques, such as the Gradient Normalization method (GradNorm) \cite{chen2018}. This method dynamically adjusts the weights $\lambda_n$ during training to ensure that the gradients from different loss components have similar magnitudes, preventing any single term from dominating the optimization process.

To train the CMM effectively with the multi-component loss defined in (\ref{eq44}), we employ an algebraic modification of the GradNorm algorithm \cite{chen2018}. 

For each loss component $\mathcal{L}^{(m)}_n$, the $L_2$ norm of its gradient with respect to the parameters of the last shared layer $w$ is computed:
\begin{align}
	G_n(t) = \lambda_n(t) \cdot \left\| \nabla_{w} L_n(t) \right\|_2 , \label{eq45}
\end{align}
where $\lambda_n(t)$ is the current weight of the $n$-th task. The average gradient norm across all active tasks is defined as $\bar{G}(t) = \mathbb{E}_n [G_n(t)]$.

To synchronize the convergence speeds, the algorithm monitors the training progress of each task relative to its initial state $\mathcal{L}^{(m)}_n(0)$:
\begin{align}
	\tilde{\mathcal{L}}^{(m)}_n(t) = \frac{\mathcal{L}^{(m)}_n(t)}{\mathcal{L}^{(m)}_n(0)}, \quad r_n(t) = \frac{\tilde{\mathcal{L}}^{(m)}_n(t)}{\mathbb{E}_j [\tilde{\mathcal{L}}^{(m)}_j(t)]}, \label{eq46}
\end{align}
where $r_n(t)$ represents the relative inverse training rate. A value of $r_n(t) > 1$ indicates that the $n$-th task is converging slower than the average, necessitating an increase in its weight $\lambda_n$.

The target gradient norm for each task is modulated by the asymmetry parameter $\alpha$:
\begin{align}
	G_n^{\text{target}}(t) = \bar{G}(t) [r_n(t)]^\alpha. \label{eq47}
\end{align}
The weights are updated by calculating the ratio between the target and current norms. To ensure numerical stability, the update factor is clamped within the range $[0.1, 10.0]$:
\begin{align}
	\lambda_n^{\text{temp}} = \lambda_n(t) \text{clamp}\left( \frac{G_n^{\text{target}}(t)}{G_n(t) + \epsilon}, 0.1, 10.0 \right). \label{eq48}
\end{align}

To prevent scale drift, the weights are renormalized such that their sum equals the number of tasks $N$. Finally, an exponential moving average with a smoothing factor $\rho$ is applied to stabilize the trajectories of $\lambda_n$ across iterations:
\begin{align}
	\hat{\lambda}_n(t+1) = \lambda_n^{\text{temp}}{N}\left/{\sum\limits_{j=1}^N \lambda_j^{\text{temp}}}\right., \\
	\lambda_n(t+1) = \rho \lambda_n(t) + (1-\rho) \hat{\lambda}_n(t+1). \label{eq_ema}
\end{align}
To adapt to different stages of the optimization process, the reference losses $\mathcal{L}^{(m)}_n(0)$ in (\ref{eq46}) are softly reset every $T_{\text{reset}}$ steps. This allows the system to recalibrate its expectations as the phase point enters the neighborhood of the stable equilibrium $\hat{\vec {y}}$, as described by the second condition (\ref{eq20}).

\subsection{Training}

The training procedure for the TRM~\cite{jolicoeurmartineau2025morerecursivereasoningtiny} serves as the primary baseline for our experimental setup. We maintained the internal tracking of supervision steps $N_{\rm super}$ as implemented in the original repository~\cite{jolicoeurmartineau2025morerecursivereasoningtiny}. Furthermore, we introduced an inner accumulation loop $N_{\rm accum}$ within each batch iteration, where $N_{\rm accum} \leq N_{\rm super}$. Notably, the case where $N_{\rm accum} = N_{\rm super}$ corresponds to the training methodology described in~\cite{wang2025hierarchicalreasoningmodel,jolicoeurmartineau2025morerecursivereasoningtiny}.

 Given the GPU memory constraints, which often prevent fitting the required number of samples into a single batch, we employed gradient accumulation. Under this approach, the model weights were updated only after every $N_{\rm grad}$ segments. For the same memory-efficiency reasons, gradient checkpointing~\cite{chen2016,gruslys2016} was utilized in several experimental runs to further reduce the peak VRAM footprint.

\subsection{Architecture}

The $L$-module and $H$-module of the TRM are represented by the same neural network (${\bm \theta}_{\rm L} \equiv {\bm \theta}_{\rm H}$ and ${\hat{\vec F}}_{\rm L} \equiv {\hat{\vec F}}_{\rm H}$)~\cite{jolicoeurmartineau2025morerecursivereasoningtiny}. This single module consists of two transformer layers (see Figure~\ref{fig1}). This architecture can be further simplified by enforcing weight sharing between these two transformer layers, effectively utilizing the same layer sequentially. We mark this modification as ``identical transformer layers''.

To mitigate overfitting and improve generalization, the training of TRM employs an exponential moving average of the weights. Another approach to prevent overfitting and simultaneously accelerate training and inference is the aggressive reduction of the total number of model parameters. Furthermore, for the CMM, which operates as a contractive mapping system, bounded nonlinear activation functions such as $\tanh$ may be preferable to unbounded functions like $\text{SiLU}$ to ensure stability and convergence.

\section{Numerical Experiments}

The performance of the proposed methods was evaluated on the Sudoku-Extreme dataset~\cite{wang2025hierarchicalreasoningmodel} as the main dataset. The neural networks and their training were implemented in the PyTorch framework~\cite{Paszke2019} (version 2.6 under CUDA 12.4).

Recognizing that many independent researchers and startups have limited access to high-performance computing clusters, efforts were focused on achieving results for the CMM that are superior or comparable to those of the TRM while operating within a restricted memory budget. Specifically, the training can be performed on hardware with 16~GB of VRAM (such as V100 or T4 GPUs) without requiring support for advanced precision formats, like \verb*|bfloat16|, and latest versions of flash attention. To accelerate the computations, automatic mixed precision (AMP) was employed. All models were trained using the Adam-Atan2 optimizer~\cite{Everett2024ScalingEA,Wang2024AdamAtan2}.

The designations in the text are as follows: $B$ is the batch size, $D$ is the hidden state size. The weights of the loss terms are as follows: $\lambda_\text{LM} = 1.0$,  $\lambda_\text{BCE} = 0.5$, $\lambda_\text{rep x}=10^3$, $\lambda_\text{rep z}=10^3$, $\lambda_\text{equil x}=1$, $\lambda_\text{equil z}=1$, $\lambda_\text{RH stable z}=10^4$, $\lambda_\text{RH unstable x}=10$.

Both training and inference were executed using compiled code via \verb*|torch.compile| (as in~\cite{wang2025hierarchicalreasoningmodel,jolicoeurmartineau2025morerecursivereasoningtiny}). It is noteworthy that models trained without such compilation demonstrate a significant degradation in accuracy. This phenomenon is attributed to the operator fusion and the utilization of Fused Multiply-Add (FMA) instructions during the compilation process, which minimize intermediate rounding errors. By performing multiple operations within GPU registers before writing back to global memory, the compiled mode preserves higher numerical precision, which is particularly critical for the stability of the iterative dynamics in our model.

\subsection{Baseline tests}

The baselines for the present study are established by comparing the proposed models against existing architectures. Tables~\ref{table1} and~\ref{table2} summarize the performance of DeepSeek R1, the original HRM, and the original TRM architectures on the Sudoku-Extreme and Maze benchmarks, respectively. The baseline data for these reference models are taken directly from the literature~\cite{wang2025hierarchicalreasoningmodel, jolicoeurmartineau2025morerecursivereasoningtiny}. To validate the experimental setup under the constrained 16~GB VRAM budget, the HRM and TRM baselines were reproduced using Automatic Mixed Precision (AMP) and gradient accumulation. While the reproduced TRM results for the Maze benchmark closely match the originally reported metrics, a noticeable performance drop is observed in the Sudoku-Extreme reproduction for both HRM and TRM. This discrepancy indicates the sensitivity of these baseline architectures to hardware-imposed training constraints, such as reduced numerical precision and limited batch configurations. Nevertheless, establishing these hardware-constrained baselines provides a fair and realistic ground for evaluating the subsequent architectural modifications.

\begin{table*}[ht!]
	\centering
	\begin{tabular}{c|c|c|c}
		\toprule
		\bf{Method} & \bf{Accuracy \%}  &  \bf{Number of} & \bf{Features}\\
		  &    &  \bf{Parameters} &  \\
		\midrule
		{\bf DeepSeek R1}~\cite{wang2025hierarchicalreasoningmodel} & 0.0 & 671B & -- \\
		{\bf HRM}~\cite{wang2025hierarchicalreasoningmodel} & 55.0 & 27M & -- \\
		{\bf TRM}~\cite{jolicoeurmartineau2025morerecursivereasoningtiny} & 87.4 & 5M & MLP-mixer \\
		{\bf TRM}~\cite{jolicoeurmartineau2025morerecursivereasoningtiny} & 74.7 & 7M & Attention  \\
		\midrule
		{\bf HRM} (our run) & 45.9  & 27M  & AMP\\
		{\bf TRM} (our run) & 79.1  & 5M  & AMP, MLP-mixer\\
		\bottomrule
	\end{tabular}
	\caption{Test accuracy on Sudoku-Extreme benchmark. Hidden size of HRM and TRM is 512. }
	\label{table1}
\end{table*}

\begin{table*}[ht!]
	\centering
	\begin{tabular}{l|c|c|l}
		\toprule
		\bf{Method} & \bf{Accuracy \%}  &  \bf{Number of} & \bf{Features}\\
		&    &  \bf{Parameters} &  \\
		\midrule
		{\bf DeepSeek R1}~\cite{wang2025hierarchicalreasoningmodel} & 0.0 & 671B & -- \\
		{\bf HRM}~\cite{wang2025hierarchicalreasoningmodel} & 74.5 & 27M & -- \\
		{\bf TRM}~\cite{jolicoeurmartineau2025morerecursivereasoningtiny} & 85.3 & 7M & Attention  \\
		\midrule
		{\bf TRM} (our run) & 84.2  & 7M  & AMP\\
		(gradient accumulation on 10 segments)& & & \\
		\bottomrule
	\end{tabular}
	\caption{Test accuracy on Maze benchmark. Hidden size of HRM and TRM is 512. }
	\label{table2}
\end{table*}

\subsection{Experiments with $\text{CMM}^{\text{NODE}}_{512}$}

Tables~\ref{table3}-\ref{table6} detail the ablation studies and optimization of the $\text{CMM}^{\text{NODE}}_{512}$ architecture. An early finding (Table~\ref{table3}) indicates that replacing the unbounded $\text{SiLU}$ activation function with the bounded $\tanh$ function can improve the stability of the latent trajectories, which naturally aligns with the requirements of a contraction mapping system. Further tuning of the high-level ($N_{\rm H}$) and low-level ($N_{\rm L}$) steps (Table~\ref{table4}) demonstrates that deeper recursive configurations yield the highest performance. Integrating the modified StableMax3 function alongside the Routh-Hurwitz stability loss terms (stable for $\hat{\vec z}_{\rm H}$ and unstable for $\hat{\vec x}$) systematically pushes the metric upwards. Table~\ref{table6} illustrates the compound effect of these modifications. Note that calculating the Routh-Hurwitz stability loss terms is a computationally expensive operation and, as our experiments show, gives an accuracy close to that of a model trained using equilibrium losses.  The introduction of the repulsion loss on the hypersphere, combined with the equilibrium loss, significantly improves the latent space representation. As a result, the $\text{CMM}^{\text{NODE}}_{512}$ reaches a substantially higher peak performance when utilizing an exponential decay learning rate scheduler. Furthermore, it is observed that StableMax5 provides performance on par with StableMax3 and slightly better than the original StableMax.

\begin{table*}[ht!]
	\centering
	\begin{tabular}{l|c}
		\toprule
	    \bf{Features} & \bf{Accuracy \%}  \\
		\midrule
		$B=256$ & 69.6 \\ 
		$\text{SiLU}$ to $\tanh$, $B=256$ & 81.8 \\
		$\text{SiLU}$ to $\tanh$, $B=64$, $N_{\rm grad}=4$,  & \\
		using the all states of the $H$-module for gradient calculation & 56.8 \\ 
		\bottomrule
	\end{tabular}
	\caption{Test accuracy of TRM on Sudoku-Extreme benchmark. General features: $D=512$, AMP, $N_{\rm super}=16$, $N_{\rm H} = 3$, $N_{\rm L} = 6$ MLP-mixer, number of parameters is 5M, $N_{\rm accum} = N_{\rm super}$. }
	\label{table3}
\end{table*}

\begin{table*}[ht!]
	\centering
	\begin{tabular}{l|l|c|l}
		\toprule
		$N_{\rm H}$ & $N_{\rm L}$ & \bf{Accuracy \%}  &  \bf{Features}  \\
		\midrule
		1 & 2 & 58.9 & $\tanh$ is activation function  \\ 
		1 & 2 & 63.8 &  \\
		1 & 4 & 68.9 &  \\
		1 & 6 & 68.7 &  \\
		2 & 1 & 64.6 &  \\
		2 & 2 & 70.5 &  \\
		2 & 4 & 75.3 &  \\
		2 & 6 & 76.4 &  \\
		2 & 4 & 78.3 & StableMax3 \\
		2 & 4 & 77.1 & StableMax3, $\tanh$ is activation function \\
		1 & 2 & 64.2 & StableMax3, RH unstable $\hat{\vec x}$ and RH stable  $\hat{\vec z}_{\rm H}$  \\
		2 & 4 & 77.9 & StableMax3, RH unstable $\hat{\vec x}$ and RH stable  $\hat{\vec z}_{\rm H}$ \\
		2 & 4 & 80.3 & StableMax3, $N_{\rm accum}=2$ \\
		3 & 6 & 84.1 & StableMax3, $N_{\rm accum}=2$ \\
		2 & 4 & 80.5 & StableMax3, $N_{\rm accum}=2$, equilibrium in $\hat{\vec x}$ and $\hat{\vec z}_{\rm H}$ \\
		3 & 6 & 84.9 & StableMax3, $N_{\rm accum}=2$, equilibrium in $\hat{\vec x}$ and $\hat{\vec z}_{\rm H}$ \\
		\bottomrule
	\end{tabular}
	\caption{Dependence of accuracy of $\text{CMM}^{\text{NODE}}_{512}$ on $N_{\rm H}$ and  $N_{\rm L}$ for Sudoku-Extreme benchmark. General features: $D=512$, AMP, $N_{\rm super}=16$, $N_{\rm grad}=1$, $B=256$, MLP-mixer, number of parameters is 5M, $\text{SiLU}$ is activation function for main part of experiments, StableMax is loss for main part of experiments.}
	\label{table4}
\end{table*}

\begin{table*}[ht!]
	\centering
	\begin{tabular}{l|c}
		\toprule
		\bf{Features} & \bf{Accuracy \%}  \\
		\midrule
		$N_{\rm accum}=3$, $N_{\rm grad}=1$ & 79.2 \\
		$N_{\rm accum}=2$, $N_{\rm grad}=1$,  & \\
		using the all states of the $H$-module for gradient calculation  & 74.6 \\
		$N_{\rm accum}=2$, $N_{\rm grad}=1$ & 78.8 \\
		$N_{\rm accum}=2$, $N_{\rm grad}=3$ & 81.8 \\
		$N_{\rm accum}=2$, $N_{\rm grad}=4$ & 81.3 \\
		$N_{\rm accum}=2$, $N_{\rm grad}=3$, repulsion for $\hat{\vec z}_{\rm H}$, & 84.1 \\
		\bottomrule
	\end{tabular}
	\caption{Test accuracy of $\text{CMM}^{\text{NODE}}_{512}$ on Sudoku-Extreme benchmark. General features: $D=512$, $B=256$, AMP, $N_{\rm super}=16$, $N_{rm H} = 2$, $N_{rm H} = 4$, StableMax3, equilibrium in $\hat{\vec x}$ and $\hat{\vec z}_{\rm H}$, MLP-mixer, number of parameters is 5M. }
	\label{table5}
\end{table*}

\begin{table*}[ht!]
	\centering
	\begin{tabular}{l|c}
		\toprule
		\bf{Features} & \bf{Accuracy \%}  \\
		\midrule
		$B=256$, $N_{\rm grad}=3$, without repulsion for $\hat{\vec x}$ & 88.0 \\
		$B=128$, $N_{\rm grad}=6$ & 89.9 \\
		$B=256$, $N_{\rm grad}=3$, $\tanh$ is activation function & 79.4 \\
		$B=256$, $N_{\rm grad}=3$ & 90.1 \\
		$B=384$, $N_{\rm grad}=3$ & 89.9 \\
		$B=250$, $N_{\rm grad}=4$ & 90.5 \\
		$B=500$, $N_{\rm grad}=2$ & 90.2 \\
		$B=250$, $N_{\rm grad}=4$, StableMax5 & 90.5 \\
		$B=250$, $N_{\rm grad}=4$, StableMax & 89.9 \\
		$B=250$, $N_{\rm grad}=4$, & \\
		identical transformer layers & \\
		(number of parameters is 2.5M) & 89.9 \\
		$B=250$, $N_{\rm grad}=4$, $D=768$,  & \\
		identical transformer layers & 88.1 \\
		$B=250$, $N_{\rm grad}=4$, & \\
		learning rate is $3\times 10^{-4}$, without equilibrium in $\hat{\vec x}$  & 75.2 \\
		$B=250$, $N_{\rm grad}=4$, & \\
		learning rate is $6\times 10^{-5}$, without equilibrium in $\hat{\vec x}$  & 91.5 \\
		$B=250$, $N_{\rm grad}=4$, without equilibrium in $\hat{\vec x}$ & \\
		scheduler: exponential decay, from $10^{-4}$ to $5 \cdot 10^{-5}$ at 30000 epochs,& \\
		$3 \cdot 10^{-5}$ from 30000 to 50000 epochs  & 91.6 \\
		\bottomrule
	\end{tabular}
	\caption{Test accuracy of $\text{CMM}^{\text{NODE}}_{512}$ on Sudoku-Extreme benchmark. General features: $D=512$, AMP, $N_{\rm super}=16$, $N_{\rm accum}=2$, $N_{\rm H} = 3$, $N_{\rm L} = 6$, StableMax3, equilibrium in $\hat{\vec x}$ and $\hat{\vec z}_{\rm H}$, repulsion for $\hat{\vec z}_{\rm H}$ and  $\hat{\vec x}$, MLP-mixer, number of parameters is 5M, learning rate is $10^{-4}$. }
	\label{table6}
\end{table*}

In Table~\ref{table6}, it is noted that the accuracy using StableMax5 is similar to the accuracy using StableMax3. The maximum accuracy of the model when learning with StableMax5 is achieved later than when training with StableMax3, and earlier than when training with StableMax.

\subsection{Experiments with $\text{CMM}^{\text{NSDE}}_{512}$}

Regarding the stochastic formulation, Tables~\ref{table7} and~\ref{table7_b} present the results for $\text{CMM}^{\text{NSDE}}_{512}$, which utilizes additive Gaussian noise ($\sigma = 0.01$). In deterministic models, rapid overfitting is a common issue due to the phase point collapsing into narrow trajectories. The injection of noise creates a robust manifold of initial states, forcing the system to learn a broader basin of attraction towards the stable equilibrium. By scaling up the gradient accumulation steps ($N_{\rm grad}=64$) and increasing the accumulation sequence in one batch ($N_{\rm accum}=16$), a highly stable optimization process is achieved. This configuration yields a significant performance improvement, setting a new peak score on the Sudoku-Extreme dataset for models of this scale. Furthermore, applying the AlgGradNorm technique and fine-tuning the epsilon parameter of Adam optimizer (to $10^{-14}$) with weight decay (Table~\ref{table7_b}) provides a highly stable training dynamic. This allows the model to consistently reach superior results without suffering from gradient domination by any single auxiliary loss term.

\begin{table*}[ht!]
	\centering
	\begin{tabular}{l|c}
		\toprule
		\bf{Features} & \bf{Accuracy \%}  \\
		\midrule
		$N_{\rm super}=8$, learning rate $3\times10^{-4}$ & 87.4 \\
		$N_{\rm super}=16$, learning rate $2\times10^{-4}$ & 91.7 \\
		$N_{\rm super}=16$, learning rate $10^{-4}$ & 91.1 \\
		$N_{\rm super}=16$, $N_{\rm accum}=16$, $N_{\rm grad}=64$ & \\
		learning rate $2\times 10^{-4}$ & \bf{93.7 }\\
		$B=1000$, $N_{\rm super}=16$, $N_{\rm accum}=16$, $N_{\rm grad}=16$ & \\
		learning rate $2\times 10^{-4}$, gradient checkpointing & 92.7 \\
		\bottomrule
	\end{tabular}
	\caption{Test accuracy of $\text{CMM}^{\text{NSDE}}_{512}$ on Sudoku-Extreme benchmark. General features: $D=512$, $B=250$, $N_{\rm grad}=4$, AMP, $N_{\rm accum}=2$, $N_{\rm H} = 3$, $N_{\rm L} = 6$, StableMax3, equilibrium in $\hat{\vec z}_{\rm H}$, repulsion for $\hat{\vec z}_{\rm H}$ and  $\hat{\vec x}$, MLP-mixer, $\sigma = 0.01$. }
	\label{table7}
\end{table*}

Training the model using AlgGradNorm is significantly faster than the baseline training, which causes the $D=512$ network to overfit quickly. In contrast, a network with $D=128$ avoids overfitting despite the accelerated training pace, owing to its substantially reduced parameter count.

\begin{table*}[ht!]
	\centering
	\begin{tabular}{l|c}
		\toprule
		\bf{Features} & \bf{Accuracy \%}  \\
		\midrule
		$N_{\rm super}=16$, $N_{\rm grad}=64$, $\lambda_{\rm LM} = 2$ & 80.2 \\
		$N_{\rm super}=2$, $N_{\rm grad}=8$, & \\
		the optimizer Adam: epsilon is $10^{-14}$, weight decay is 2  & 89.8 \\
		$N_{\rm super}=3$, $N_{\rm grad}=12$, & \\
		the optimizer Adam: epsilon is $10^{-14}$, weight decay is 2  & 91.7 \\
		\bottomrule
	\end{tabular}
	\caption{Test accuracy of $\text{CMM}^{\text{NSDE}}_{512}$ on Sudoku-Extreme benchmark with gradient normalization (AlgGradNorm). General features: $D=512$, $B=250$, AMP, $N_{\rm accum}=16$, $N_{\rm H} = 3$, $N_{\rm L} = 6$, StableMax3, equilibrium in $\hat{\vec z}_{\rm H}$, repulsion for $\hat{\vec z}_{\rm H}$ and  $\hat{\vec x}$, MLP-mixer, $\sigma = 0.01$, weight decay is 1, learning rate $2\times10^{-4}$, . }
	\label{table7_b}
\end{table*}

\subsection{Experiments with $\text{CMM}^{\text{NODE}}_{128}$ and  $\text{CMM}^{\text{NSDE}}_{128}$}

Tables~\ref{table8} and~\ref{table9} explore the extreme limits of parameter efficiency. By aggressively reducing the hidden dimension to $D=128$, the baseline TRM and standard $\text{CMM}^{\text{NODE}}$ models struggle severely, experiencing a drastic drop in performance due to the lack of expressivity in the tiny 0.52M parameter space. However, the application of the NSDE framework enables a dramatic recovery. By increasing the batch size to $B=1000$ to stabilize the noise estimation, employing AlgGradNorm to balance the complex multi-component loss, and freezing the input embeddings after 2500 epochs, the predictive capability is largely restored. Remarkably, when weight-sharing is enforced across the recursive steps (identical transformer layers), the parameter count is halved to an ultra-tiny 0.26M parameters. Adjusting the Adam epsilon parameter to an extremely small value ($10^{-14}$) to accommodate the micro-adjustments in the gradients allows this miniature $\text{CMM}^{\text{NSDE}}_{128}$ model to reach a highly competitive performance level (Table~\ref{table9}). This demonstrates that enforcing a mathematically rigorous contraction mapping allows a model with merely a quarter of a million parameters to rival the performance of baseline architectures that are substantially larger.

\begin{table*}[ht!]
	\centering
	\begin{tabular}{l|l|l|c|c}
		\toprule
		$N_{\rm H}$ & $N_{\rm L}$  &  \bf{Features} & \bf{Accuracy \%} & \bf{Number of} \\
		& & & & \bf{Parameters}  \\
		\midrule
		2 & 4 & $\text{CMM}^{\text{NODE}}_{128}$, $B=256$ & 20.3 & 0.52M \\
		3 & 6 & $\text{TRM}$, $B=250$, $N_{\rm grad}=4$, $N_{\rm super}=8$, Attention & 23.8 & 0.52M \\
		3 & 6 & $\text{CMM}^{\text{NODE}}_{128}$, $B=250$, $N_{\rm grad}=4$, & & \\
		  &   &  learning rate is $3{\times}10^{-4}$, Attention & 31.4 & 0.52M \\
		2 & 4 & $\text{CMM}^{\text{NSDE}}_{128}$, $B=250$ & & \\
		  &   & $N_{\rm grad}=4$, $N_{\rm super}=8$ & 63.4 & 0.52M \\
		3 & 6 & $\text{CMM}^{\text{NSDE}}_{128}$, $B=250$ & & \\
		  &   & $N_{\rm grad}=4$, $N_{\rm super}=8$, learning rate is $3{\times}10^{-4}$ & 71.4 & 0.52M \\
		3 & 6 & $\text{CMM}^{\text{NSDE}}_{128}$, $B=1000$ & & \\
		  &   & $N_{\rm grad}=1$, $N_{\rm super}=8$, learning rate is $3{\times}10^{-4}$ & 74.4 & 0.52M \\
 		3 & 6 & $\text{CMM}^{\text{NSDE}}_{128}$, $B=250$, $N_{\rm accum}=8$, & & \\
 		&   & $N_{\rm grad}=32$, $N_{\rm super}=8$, learning rate is $10^{-4}$ & 34.8 & 0.52M \\
 		3 & 6 & $\text{CMM}^{\text{NSDE}}_{128}$, $B=250$, $N_{\rm accum}=16$, & & \\
 		&   & $N_{\rm grad}=64$, $N_{\rm super}=16$, learning rate is $2\times10^{-3}$ & 70.9 & 0.52M \\
 		3 & 6 & $\text{CMM}^{\text{NSDE}}_{128}$, $B=250$, $N_{\rm accum}=4$, & & \\
 		&   & $N_{\rm grad}=16$, $N_{\rm super}=4$, learning rate is $2\times10^{-3}$ & 61.0 & 0.52M \\
 		3 & 6 & $\text{CMM}^{\text{NSDE}}_{128}$, $B=250$, $N_{\rm accum}=4$, & & \\
 		  &   & $N_{\rm grad}=16$, $N_{\rm super}=4$, learning rate is $2\times10^{-3}$ & &\\
 		  &   & freeze embedding after 2500 epoch & 61.5 & 0.52M \\
	    3 & 6 & $\text{CMM}^{\text{NSDE}}_{128}$, $B=1000$, $N_{\rm accum}=16$, & & \\
	 	  &   & $N_{\rm grad}=16$, $N_{\rm super}=16$, learning rate is $2\times10^{-3}$ & &\\
	 	  &   & freeze embedding after 2500 epoch, AlgGradNorm,  & \bf{85.0} & 0.52M \\
	 	3 & 6 & $\text{CMM}^{\text{NSDE}}_{128}$, $B=1000$, $N_{\rm accum}=16$, & & \\
	 	  &   & $N_{\rm grad}=16$, $N_{\rm super}=16$, learning rate is $2\times10^{-3}$ & &\\
	 	  &   & freeze embedding after 2500 epoch, AlgGradNorm, & & \\
	 	  &   & identical transformer layers  & \bf{84.7} & \bf{0.26M} \\
		\bottomrule
	\end{tabular}
	\caption{Dependence of accuracy of $\text{CMM}$ on $N_{\rm H}$ and  $N_{\rm L}$ for Sudoku-Extreme benchmark. General features: $D=128$, AMP, $N_{\rm super}=16$, $N_{\rm accum}=2$, $N_{\rm grad}=3$, MLP-mixer, number of parameters is 0.5M, StableMax3 is the loss function for the majority of the experiments, equilibrium in $\hat{\vec x}$ and $\hat{\vec z}_{\rm H}$, repulsion for $\hat{\vec z}_{\rm H}$ and  $\hat{\vec x}$, learning rate is $10^{-4}$, epsilon of the optimizer Adam is $10^{-12}$, $\hat{\vec z}_{\rm H}$ and $\hat{\vec z}_{\rm L}$ are initialized by $\hat{\vec x}$ and $\hat{\vec 0}$, respectively, at the beginning of the 1st segment of training.}
	\label{table8}
\end{table*}

\begin{table*}[ht!]
	\centering
	\begin{tabular}{l|c|c}
		\toprule
		\bf{Features} & \bf{Accuracy \%} & \bf{Number of} \\
		& & \bf{Parameters}  \\
		\midrule
		initialization of $\hat{\vec z}_{\rm H}$ and $\hat{\vec z}_{\rm L}$ as in baseline, epsAdam =$ 10^{-8}$ & 77.8 & 0.52M  \\
		epsAdam = $10^{-8}$ & 81.8 & 0.52M \\
		epsAdam = $10^{-12}$ & 85.0 & 0.52M\\
		epsAdam = $10^{-14}$ & \bf{86.1} & \bf{0.52M}\\
		epsAdam = $10^{-14}$, identical transformer layers  & \bf{85.4} & \bf{0.26M} \\
		\bottomrule
	\end{tabular}
	\caption{Dependence of accuracy of \text{CMM} on different features for Sudoku-Extreme benchmark. General features: $D=128$, $N_{\rm H}=3$, $N_{\rm L}=6$, AMP, $N_{\rm super}=16$, $N_{\rm accum}=16$, $N_{\rm grad}=16$, $B=1000$, MLP-mixer, number of parameters is 0.5M, StableMax3 is the loss function for the majority of the experiments, equilibrium in $\hat{\vec x}$ and $\hat{\vec z}_{\rm H}$, repulsion for $\hat{\vec z}_{\rm H}$ and  $\hat{\vec x}$, learning rate is $2\times10^{-3}$, freeze embedding after 2500 epoch, AlgGradNorm, epsilon of the optimizer Adam is epsAdam = $10^{-12}$, $\hat{\vec z}_{\rm H}$ and $\hat{\vec z}_{\rm L}$ are initialized by $\hat{\vec x}$ and $\hat{\vec 0}$, respectively, at the beginning of the 1st segment of training.}
	\label{table9}
\end{table*}

To verify that the proposed architectural enhancements and loss formulations are not restricted to the Sudoku dataset, the ultra-tiny $\text{CMM}$ architecture is also evaluated against the baselines established in Table~\ref{table2} on the Maze benchmark (Table~\ref{table11}). Under the highly constrained setting ($D=128$, identical transformer layers, 0.26M parameters, and Adam epsilon $=10^{-14}$), the model successfully converges and maintains strong predictive performance. This result confirms that the NSDE framework, and the equilibrium and repulsion loss penalties provide robust, task-agnostic regularization that significantly enhances the algorithmic reasoning capabilities of extreme small-scale neural networks.

\begin{table*}[ht!]
	\centering
	\begin{tabular}{l|c|c}
		\toprule
		\bf{Features} & \bf{Accuracy \%} & \bf{Number of} \\
		& & \bf{Parameters}  \\
		\midrule
		$\lambda_{\rm LM} = 10$, epsAdam = $10^{-14}$, identical transformer layers  & \bf{82.2} & \bf{0.26M} \\
		\bottomrule
	\end{tabular}
	\caption{Dependence of accuracy of $\text{CMM}$ on different features for Maze benchmark. General features: $D=128$, $N_{\rm H}=3$, $N_{\rm L}=4$, AMP, $N_{\rm super}=16$, $N_{\rm accum}=4$, $N_{\rm grad}=4$, $B=100$, MLP-mixer, StableMax3 is the loss function for the majority of the experiments, equilibrium in $\hat{\vec x}$ and $\hat{\vec z}_{\rm H}$, repulsion for $\hat{\vec z}_{\rm H}$ and  $\hat{\vec x}$, learning rate is $2\times10^{-4}$, freeze embedding after 5000 epoch, AlgGradNorm, epsilon of the optimizer Adam is epsAdam = $10^{-12}$, $\hat{\vec z}_{\rm H}$ and $\hat{\vec z}_{\rm L}$ are initialized by $\hat{\vec x}$ and $\hat{\vec 0}$, respectively, at the beginning of the 1st segment of training.}
	\label{table11}
\end{table*}

\newpage
\section{Conclusions}

In this paper, we introduced the Contraction Mapping Model (CMM), a novel architecture that bridges the gap between compact recursive reasoning models and the rigorous mathematical framework of continuous dynamical systems. By reformulating the discrete iterative steps of baseline hierarchical models into Neural Ordinary and Stochastic Differential Equations, we provided a principled approach to stabilizing latent reasoning dynamics. We demonstrated that enforcing contraction mapping properties, by using equilibrium constraints and the Routh-Hurwitz stability criterion, effectively guides the phase point of the system toward a stable, correct solution. Furthermore, the integration of a hyperspherical repulsion loss successfully mitigates representational collapse, while the use of adaptive loss balancing via AlgGradNorm and polynomial StableMax approximations ensures robust multi-objective optimization.

Our empirical results on complex algorithmic benchmarks, specifically Sudoku-Extreme and Maze, underscore the profound advantages of this approach. The injection of noise within the NSDE framework not only prevented rapid overfitting but also allowed the model to construct a resilient, broad basin of attraction in the latent space. In particular, we achieved highly accurate predictions while aggressively compressing our model to just 0.26 million parameters, a significant reduction compared to existing approaches, without sacrificing performance.

These findings suggest that massive parameter scaling is not a strict prerequisite for complex algorithmic reasoning if the underlying latent dynamics are mathematically constrained and properly optimized. Future work will explore the application of CMMs to broader, open-ended reasoning benchmarks such as ARC-AGI, as well as the potential integration of these ultra-efficient, mathematically rigorous reasoning engines into resource-constrained edge devices and larger foundational models.

\newpage
\bibliographystyle{IEEEtran}
\bibliography{Eskin_CMM_library}

\end{document}